\documentclass{article} 
\usepackage{iclr2027_conference,times}

\usepackage{amsmath,amsfonts,bm}

\def\eqref#1{equation~\ref{#1}}

\def\1{\bm{1}}

\DeclareMathAlphabet{\mathsfit}{\encodingdefault}{\sfdefault}{m}{sl}
\SetMathAlphabet{\mathsfit}{bold}{\encodingdefault}{\sfdefault}{bx}{n}

\usepackage{hyperref}
\usepackage{url}

\usepackage{graphicx}

\newcommand{\methodname}{MemUse}

\title{Delivered, Not Used: \\
Evaluating Agent Memory at Its Point of Use}

\author{Changran Xu, Zeju Li, Qiang Xu\\
Department of Computer Science and Engineering\\
The Chinese University of Hong Kong\\
\texttt{\{crxu25,zjli24,qxu\}@cse.cuhk.edu.hk}
}

\iclrfinalcopy 
\begin{document}

\maketitle

\begin{abstract}
Task success, retrieval quality, and cost do not show whether an LLM agent used a delivered memory or what followed from that use. We introduce \methodname{}, which evaluates memory at its point of use in recorded trajectories: each memory delivery is compared with what the agent could already see and followed through the agent's later actions and their consequences. An agentic judge rates each delivered entry's relevance, accuracy, novelty, influence, and observed effect; programmatic checks enforce complete reading of cited evidence, and fixed rules aggregate the ratings into verdicts linked to memory histories and delivery cost. On 8,000 SWE-bench Verified episodes from three memory systems, resolve rates stay within 0.4 percentage points of a no-memory baseline, yet \methodname{} shows that most delivered entries were irrelevant, most relevant entries went unused, and in two systems most deliveries restated content the agent could still see. It also surfaces effects that outcomes hide: 23.2\% of episodes with beneficial memory use still failed, and 5 of 16 with harmful use still succeeded.
\end{abstract}

\section{Introduction}
\label{sec:introduction}

As large language models (LLMs) become more capable, agents built on them are increasingly deployed on long-running tasks, such as repository-level software engineering~\citep{jimenez2024swebench,huang2026deepswemeasuringfrontiercoding} and multi-step workflows in terminal environments~\citep{merrill2026terminalbench,desai2026swemarathonagentsautonomouslycomplete}. These tasks can accumulate more information than an agent's finite context window can hold, and even within this limit, an agent's performance can decline as the context grows~\citep{du-etal-2025-context,xia2026diagnosingmitigatingcontextrot}. Earlier decisions, codebase knowledge, and lessons from previous attempts often remain useful within and across tasks. Memory systems, such as MemGPT~\citep{packer2024memgptllmsoperatingsystems},  Mem0~\citep{chhikara2025mem0buildingproductionreadyai} and MemGovern~\citep{wang2026memgovernenhancingcodeagents}, preserve and organize this information for repeated delivery during agent execution. Delivery, however, does not guarantee that the agent uses the memory or benefits from it, and delivered memory occupies part of the same finite context whose growth can degrade performance. This raises a central question: \emph{when does delivered memory help the agent, remain unused, or misdirect it?}

Existing memory evaluations provide only a partial answer, especially for agents performing complex, long-running tasks. 
Benchmarks such as LongMemEval~\citep{wu2025longmemeval} and LoCoMo~\citep{maharana-etal-2024-evaluating} primarily assess memory through question answering over conversation histories, with LoCoMo additionally evaluating event summarization and dialogue generation.
However, these benchmarks do not directly assess how retrieved information guides agents' actions during task execution.
Moving beyond conversational QA, VibeMemBench~\citep{fan2026vibemembenchevaluatingmemorysystems} evaluates memory systems on repository tasks with executable tests, comparing memory-equipped agents with a matched memory-off baseline. Such comparisons measure differences in overall task performance, but not how individual memory deliveries influence agent behavior during execution. Useful guidance may be followed even when the agent ultimately fails a task, while harmful guidance may be corrected before it succeeds. Outcomes also cannot explain null results: when memory systems fail to beat the memory-off baseline, as in eleven of twelve VibeMemBench pairings, it remains unclear whether the delivered memory was redundant, ignored, or helpful at some steps but harmful at others. Moreover, apparent memory gains can be inflated by evaluation artifacts~\citep{son2026persistentmemorymultiagentllm}.

Recent studies diagnose memory at a finer grain, evaluating memory operations, failures, and use~\citep{hao2026memopsbenchmarkinglifecyclememory,deng2026memtracetracingattributingerrors,cao2026memcalibbenchmarkingoptimizingmemory,liu2026evalmemoperationleveldiagnosticframework}. Behavioral analyses further examine whether retrieved facts appear in executed tool-call arguments and how memory-induced actions persist or are corrected~\citep{mishra2026doesmemoryhelpcostaware,chen2026compliancetrapdiagnosingai}. These diagnostics rely on gold answers, reference operation traces, annotated use targets, known fact values, or controlled memory interventions, so applying them to an existing agent workflow requires reference information or additional executions beyond its execution record.
Meanwhile, reference-free trajectory evaluators such as TRACE~\citep{pmlr-v306-kim26bg} avoid this requirement by assessing reasoning and tool use against accumulated execution evidence, but they do not explicitly distinguish the availability of specific information from evidence of its use and subsequent consequences.
These studies leave open how to evaluate individual memories dynamically delivered during agent execution, without reference answers or controlled memory interventions.

Evaluating memory from execution records alone raises three challenges. First, the same entry may be delivered many times and in different forms, such as its full text or a search preview, so an evaluation must identify exactly what the agent saw. Second, the agent may already have the same information in its visible context, so an action that matches the memory does not by itself show memory use. Third, memory may shape actions many steps later, and those actions may be corrected or reversed, so its effects must be followed through the rest of the trajectory.

\begin{figure}[ht]
    \centering
    \includegraphics[width=0.9\linewidth]{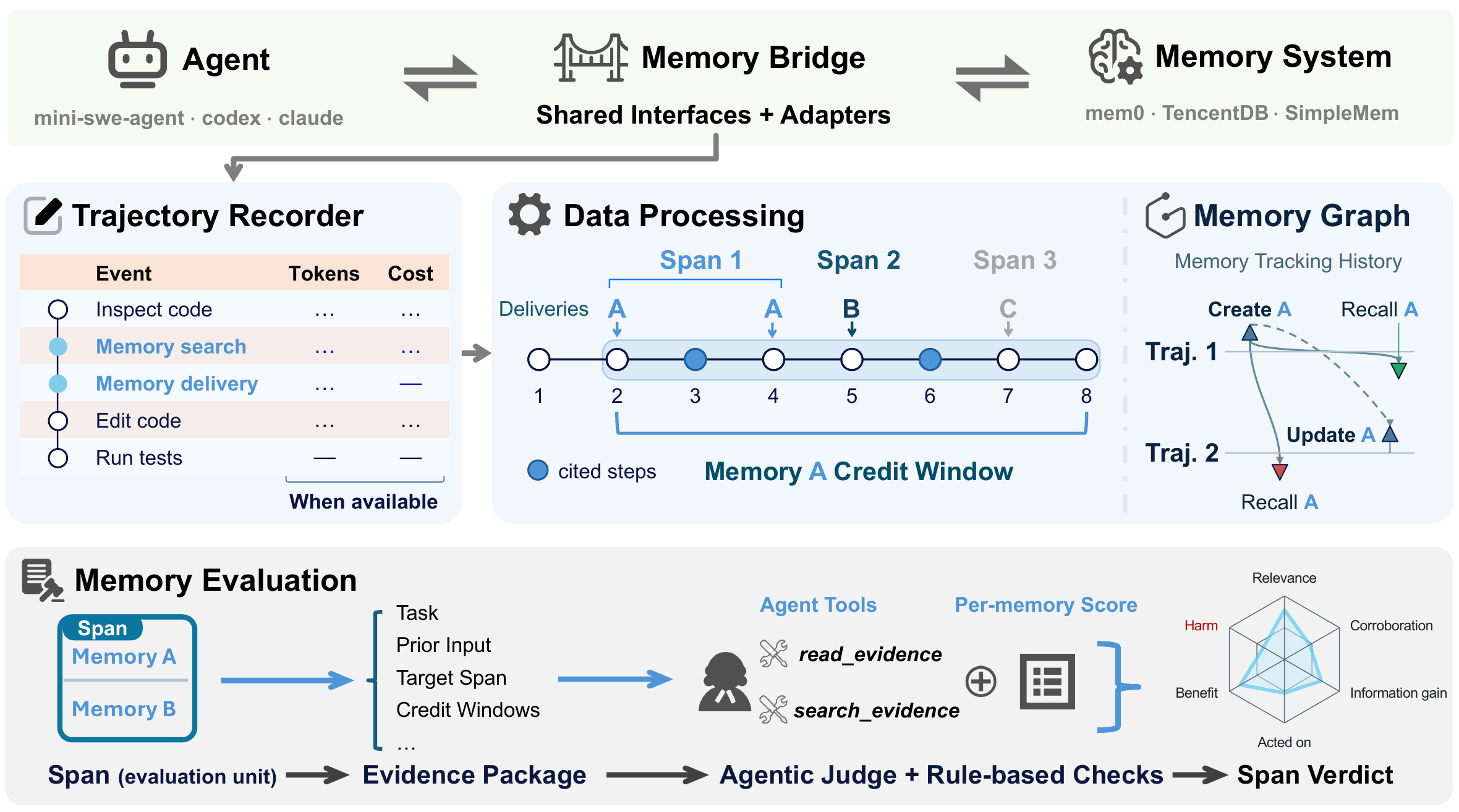}
    \caption{Overview of \methodname{}. The memory bridge connects agents to memory systems and annotates memory operations.
The trajectory recorder captures model calls, with token usage and cost when available.
Within each episode, data processing groups identical repeated deliveries into injection spans while earlier copies remain visible and assigns each entry a credit window for attributing actions.
The memory graph links deliveries to recorded entry histories across trajectories.
For each eligible span, an agentic judge assesses each entry's information quality, use, and observed effects from its exposed text, prior context, and subsequent trajectory.
Programmatic checks enforce evidence coverage and credit windows, and fixed rules derive the span verdict.}
    \label{fig:memeval-overview}
\end{figure}

To address these challenges, we introduce \textbf{\methodname{}}, a framework for evaluating individual memory deliveries in recorded agent trajectories as shown in Figure~\ref{fig:memeval-overview}. Our central idea is to judge each memory delivery in context, rather than assessing stored entries in isolation.
For each delivery, \methodname{} identifies exactly which text the agent received, compares that text with what the agent already knew, and follows what the agent did next through the rest of the trajectory. This separates the quality and novelty of the delivered information from evidence that the agent used it and from what that use led to.
To our knowledge, \methodname{} is the first reference-free framework to jointly assess the use and observed effects of individual memory deliveries without a dedicated memory benchmark, reference solution, task label, or additional execution.

To make these assessments verifiable, an agentic judge supports its assessments with cited trajectory evidence. Programmatic checks enforce evidence-access and coverage requirements, while fixed rules produce verdicts and retain uncertainty when evidence is insufficient. Linking these verdicts to memory histories and delivery costs helps identify irrelevant or redundant content, useful guidance that remains unused, and misleading guidance associated with unproductive actions. A memory bridge and a pass-through trajectory recorder support these assessments across agent harnesses and memory systems, covering both memory injected by the host and memory retrieved by the agent. 

We examine what outcome comparisons miss by applying \methodname{} to 8,000 recorded episodes, each a single agent execution of a task. These episodes cover all 500 tasks of SWE-bench Verified~\citep{jimenez2024swebench} and four configurations of three memory systems: MemGovern~\citep{wang2026memgovernenhancingcodeagents}, Mem0~\citep{chhikara2025mem0buildingproductionreadyai}, and TencentDB Agent Memory~\citep{tencentcloud2026tencentdbagentmemory}. In these evaluations, the judge also receives each task's gold patch through the optional reference-solution mode. Task outcomes barely separate the four configurations, since their resolve rates lie within 0.4 percentage points of a no-memory baseline, less than the spread across the baseline's own runs. \methodname{} nonetheless separates the systems and reveals patterns consistent with these small outcome differences.
Most delivered entries were irrelevant, 51--80\% of the relevant ones remained unused in each system, and the entries that the agent acted on helped it far more often than they misdirected it. Task outcomes also concealed these effects, since 23.2\% of episodes with beneficial memory use still failed, and 31.3\% of those with harmful use still succeeded.

\section{Related Work}
\label{sec:related-work}

\paragraph{Memory systems for LLM agents.}
Memory systems for LLM agents differ in how they represent, organize, and reuse past information. Generative Agents~\citep{Park2023GenerativeAgents} and Reflexion~\citep{shinn2023reflexion} use stored observations or verbal reflections to guide later behavior. Others focus on memory organization: MemGPT~\citep{packer2024memgptllmsoperatingsystems} manages information across memory tiers, Mem0~\citep{chhikara2025mem0buildingproductionreadyai} extracts and updates conversational facts, and EverMemOS~\citep{hu-etal-2026-evermemos} and Hindsight~\citep{latimer-etal-2026-hindsight} organize interaction histories into structured long-term memory. Beyond conversational facts, MemGovern~\citep{wang2026memgovernenhancingcodeagents} distills GitHub issues, pull requests, and patches into reusable experience cards, and TencentDB Agent Memory~\citep{tencentcloud2026tencentdbagentmemory} shares skills and knowledge across agents. Complementing these efforts, \methodname{} evaluates whether and how individual memory deliveries guide agents' subsequent actions.

\paragraph{Memory evaluation benchmarks.}
Many memory benchmarks evaluate memory by end-task performance. LongMemEval~\citep{wu2025longmemeval} tests chat assistants on abilities such as knowledge updates and abstention, LoCoMo~\citep{maharana-etal-2024-evaluating} probes temporal and causal reasoning over long, multi-session conversations, and MemoryAgentBench~\citep{hu2026evaluating} uses incremental interactions to evaluate retrieval, test-time learning, long-range understanding, and selective forgetting. For executable tasks, VibeMemBench~\citep{fan2026vibemembenchevaluatingmemorysystems} evaluates coding agents with and without memory on repository tasks where prior experience was observed to improve performance in a reference setting. Its main memory-system comparison provides one top-ranked retrieved experience before execution and attributes underperforming cases to coverage misses, ranking misses, form degradation, or use failures. Such comparisons must also control for leakage, confounds, and run-to-run noise~\citep{son2026persistentmemorymultiagentllm}. \methodname{} evaluates memory delivered during execution, whether injected by the host or retrieved by the agent, and judges each delivery through recorded actions and consequences, so it can surface useful guidance in failed runs and harmful detours in successful ones.

\paragraph{Fine-grained memory diagnosis.}
Diagnostic benchmarks localize memory failures that aggregate scores hide. MemOps~\citep{hao2026memopsbenchmarkinglifecyclememory} evaluates lifecycle operations against gold operation traces, and MemTrace~\citep{deng2026memtracetracingattributingerrors} uses gold answers to trace failed cases back to decisive faulty operations. EvalMem~\citep{liu2026evalmemoperationleveldiagnosticframework} separates encoding, retrieval, and generation defects using reference facts and oracle evidence, and MemCalib~\citep{cao2026memcalibbenchmarkingoptimizingmemory} measures proposition-level over-use and under-use against annotated target levels. Two studies examine agent actions more directly. MERIT~\citep{mishra2026doesmemoryhelpcostaware} measures how often agents act on a correctly retrieved fact in their tool calls, and Entry--Propagation--Recovery (E-P-R)~\citep{chen2026compliancetrapdiagnosingai} compares paired trajectories under varied memory content and injection schedules to locate where memory first diverts an action, whether the diversion persists, and whether the agent recovers. Each of these diagnoses relies on ground truth prepared in advance or on additional controlled runs. \methodname{} works from the recorded run alone and evaluates naturally delivered memories from recorded execution evidence.

\paragraph{Trajectory-level agent evaluation.}
Trajectory evaluators provide a broader basis for examining agent behavior. TRACE~\citep{pmlr-v306-kim26bg} scores efficiency, hallucination, and adaptivity against an evidence bank of the agent's earlier tool calls and observations, with no reference trajectory. TRAIL~\citep{deshpande2025trailtracereasoningagentic} benchmarks error identification and localization, Who\&When~\citep{zhang2025which} benchmarks the localization of errors and failure-responsible steps in agent traces, and TrajDebug~\citep{qi2026trajdebugtracingerrorlifecycle} traces errors through their resolution and terminal impact. Agent-as-a-Judge~\citep{pmlr-v267-zhuge25a} extends LLM-as-a-judge~\citep{3666122.3668142} with artifact inspection and evidence retrieval to check task requirements. Attributing behavior to a specific memory delivery requires an additional distinction: an action consistent with retrieved guidance may already have been planned before that delivery. \methodname{} adapts agentic inspection to this attribution problem. Its judge compares exactly what each delivery exposed with the agent's earlier context, and programmatic checks confirm that every judgment cites fully read trajectory steps and covers the steps it requires, making each attribution verifiable against the recorded trajectory.

\paragraph{Memory valuation and credit assignment.}
Another line of work scores individual memory entries to guide memory management. MemLens~\citep{10.14778/3827998.3828077} approximates each memory unit's Shapley-style contribution to a lightweight proxy model's judged response quality and retains high-value units. Hindsight Memory-PRM~\citep{jia2026hindsightmemoryprmsupervisingmemory} assigns entry-level credit from retrieval logs, answer citations, and controlled deletion-and-reanswer probes, then propagates it along version chains as a proxy reward for training a memory manager. Causal Memory Policy~\citep{behnam2026causalmemorypolicymaking} randomizes retrieval exposure with known propensities to estimate utility for memories in an exposure pool, including entries missed by the original retriever. All three measure an entry's contribution through scored answers under deliberately altered memory contexts (sampled subsets, targeted deletions, or randomized exposure). \methodname{} instead gives an observational account of realized use. For each delivered entry, it reports whether and how the agent used the entry and what followed, linked to its history in the memory graph and the estimated size of its delivered text.
\section{Methodology}
\label{sec:method}


\methodname{} evaluates memory at its point of use. Its unit of evaluation is a \emph{memory delivery in context}. This unit combines a delivered memory entry with the context that the agent already had, the actions that the agent took next, and the consequences of those actions. Figure~\ref{fig:memeval-overview} outlines how \methodname{} builds and judges these units. We first record memory operations and model calls, and we reconstruct each delivery in context from these records. An agentic judge then assesses the entries of eligible spans. Programmatic checks verify that the judge read the required evidence, and fixed rules convert the checked assessments into span verdicts. Finally, the memory graph links each verdict to memory histories and resource use.

\subsection{Memory Delivery in Context}
\label{subsec:formulation}

An agent solves a task $q$ through interleaved model calls and tool interactions~\citep{yao2023react}. An \emph{episode} is one execution of the agent on a task, represented by a trajectory $\tau=(s_1,\dots,s_T)$. Each step $s_t=(x_t,y_t,o_t)$ contains one completed model call with its input $x_t$, its response $y_t$, and the resulting tool observations $o_t$. Memory reaches a model input either through \emph{host-scheduled} insertion or as the result of an \emph{agent-directed} memory-tool call. A delivery $d$ places an ordered list of entries $M_d$ in a model input. Its \emph{consuming step} $t_d$ is the first step whose model input contains $d$, and $x_{t_d}$ is its \emph{consuming input}. An entry takes the form that its memory system delivers, such as a stored record, a search preview, an index, or a synthesized answer. Each entry belongs to a memory \emph{item}, so different deliveries can show the same item in different versions or forms. The \emph{exposed text} $\epsilon_d(m)$ is the part of entry $m$ that appears in the consuming input. The \emph{exposure} of $m$ is \emph{full} when the consuming input contains the whole entry and \emph{clipped} when it contains only a prefix, for example after a length limit truncates the entry. Exposure is \emph{unknown} when the recording verifies neither, and $\epsilon_d(m)$ is then undetermined.

We evaluate each entry as a \emph{memory delivery in context},
\begin{equation}
\label{eq:delivery-in-context}
c_d(m)=\bigl(q,\; B_d,\; \epsilon_d(m),\; W_d(m),\; \tau\bigr),
\end{equation}
where the \emph{prior context} $B_d$ is the consuming input $x_{t_d}$ without the text of delivery $d$, and the \emph{credit window} $W_d(m)\subseteq\{t_d,\dots,T\}$ contains the steps whose actions can be credited to entry $m$. The exposed text limits the evaluation to the content that the agent received. The prior context holds everything else that the agent could see at that call, including earlier messages, tool results, and other memory deliveries. It is the reference for relevance and novelty. Together with earlier steps, it also separates memory use from actions that prior information or plans explain. Later steps show consequences, including corrections and reversals.

We characterize each entry along three separate aspects. \emph{Information} comprises its relevance, accuracy, and novelty; \emph{use} is its influence on actions; and \emph{consequence} is the observed effect of that use. Repeated identical deliveries form \emph{injection spans}, and each entry receives one assessment per span. Fixed rules aggregate these assessments into a span verdict, which pairs a triage label with an entry-share profile. The verdict describes observed behavior in the recorded episode, not a counterfactual effect on task success. The default evaluation needs neither a reference solution nor a reference trajectory.

\subsection{Recording Memory Deliveries}
\label{subsec:recording}

Evaluating a delivery requires two records: which input text came from memory, and what the agent actually received. The memory bridge identifies memory text through adapters for agent harnesses and memory systems. It annotates memory creation, updates, searches, and deliveries. For each delivery, it records every \emph{placement}, a piece of memory text at one location in the model request. A \emph{placement proof}, such as a digest of the carrying message or a path into the request, identifies this location. For memory systems that are built into a harness or run behind their own proxy, an observer emits the same annotations.

A pass-through trajectory recorder captures what the agent actually received. It records requests and responses at the model-API boundary, including memory that a harness inserts into an outbound request but does not save in its history. The recorder normalizes calls into a common event log, keeps the model calls of the memory system out of the agent-step sequence, and retains reported token usage and cost when available. It checks each placement proof against the recorded request and links the delivery to the model call that carried it, the \emph{consuming call}. When this call completed an agent step, that step is the \emph{attributed consuming step} of the delivery. The recorder also logs when the harness removes or rewrites earlier messages, which shows when memory text leaves the input of the agent.

\subsection{Constructing the Evaluation Context}
\label{subsec:injection-spans}

From these records, we construct the evaluation context of Eq.~(\ref{eq:delivery-in-context}) for each delivered entry. We recover the exposed text and the prior context, group repeated deliveries into injection spans that are judged once, assign a credit window to each entry of a span, and decide which spans the judge receives.

\paragraph{Exposed text and prior context.}
Within each episode, we match each delivery to its verified consuming call and compare each entry with the recorded placement text to decide its exposure. The judge always sees the placement text. For full or clipped exposure, it also receives the exposed text of the entry. For unknown exposure, it does not receive the stored value. Type labels and confidence scores that exist only in the annotations are also withheld, because the agent never saw them. We reconstruct $B_d$ from the recorded consuming input by removing the text of delivery $d$. The prior context therefore contains only text that is still present at step $t_d$. An entry delivered earlier but no longer in the input is not part of $B_d$. Steps that a context rewrite removed are also not part of $B_d$, but the judge can still read them to check what the agent did or planned before the delivery.

\paragraph{Injection spans.}
Memory is often delivered repeatedly, for example when a host inserts the same entries before every model call. An \emph{injection span} groups such repeats into one unit. A delivery joins an existing span only under two conditions. First, it repeats the entries of the span exactly, with the same order, the same placements, and the same exposure. Second, an earlier copy is still visible in its consuming input. Otherwise, the delivery starts a new span. The judge assesses each eligible span once. The later deliveries of a span are \emph{restatements}. They add no verdicts, but every copy contributes to delivered text volume. For a span $S$, $M_S$ denotes its entries, $B_S$ the prior context at its first consuming step, and $W_S(m)$ the credit window of entry $m$. All deliveries in $S$ share the same exposure, so $\epsilon_S(m)$ denotes the exposed text of entry $m$.

\begin{figure}[t]
    \centering
    \includegraphics[width=0.9\linewidth]{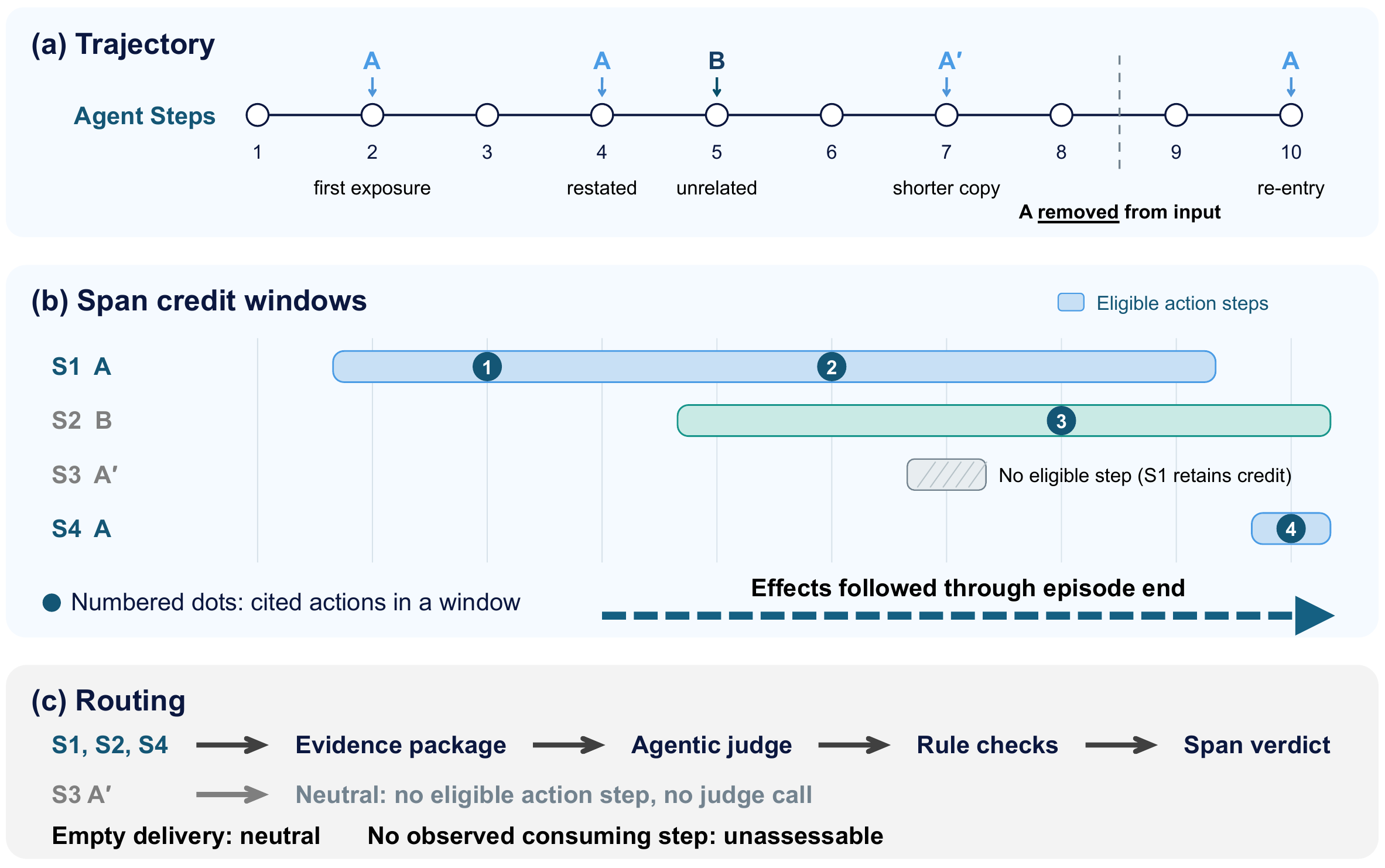}
    \caption{Credit windows and judge routing. (a) A's deliveries at steps 2 and 4 share $S_1$; re-entry after removal starts $S_4$ with a fresh prior context. (b) The shorter copy $A'$ adds no text or identifier beyond the visible $A$, so $A'$ in $S_3$ receives an empty window and $A$ in $S_1$ retains credit through step 9. Effects of credited actions are followed through the episode end. (c) $S_1$, $S_2$, and $S_4$ are judged; $S_3$ receives a rule-derived neutral triage label without a judge call. Empty deliveries receive a neutral placeholder and are excluded from triage rates.}
    \label{fig:credit-windows}
\end{figure}

\paragraph{Credit windows.}
Credit windows divide the action credit for an item among the spans that deliver it. The window $W_S(m)$ starts at the first consuming step of $S$. It ends when an entry of the same item in a later span takes over the credit, or at the end of the episode if no entry does. For verified exposure, a later entry can take over when its text equals or contains $\epsilon_S(m)$. It must add text or an identifier, arrive when the earlier copy is no longer provably visible, or be a separately recorded full read reached through an explicit selection. Otherwise, a later copy that adds nothing to still-visible content receives an empty window, and $W_S(m)$ continues. When either exposure is unknown, the boundary rule uses item identity. A window only limits which actions an entry can claim. A credited action must occur inside the window, but its consequences are followed to the end of the episode. A test run after the window closes can therefore still show that an action inside the window helped or harmed. Removal of the entry text from the input does not close the window, because the agent can still act on guidance that it read earlier. Figure~\ref{fig:credit-windows} shows retained credit and a re-entry with a new prior context. Appendix~\ref{app:window-edge-cases} gives the exact boundary rule, including unknown exposure, later copies with narrower or different text, and copies consumed in the same step.

\paragraph{Eligibility.}
The judge receives a span only if three conditions hold. The span has at least one entry and an attributed consuming step, at least one of its entries has a nonempty credit window, and its episode passes the recording checks below. A delivery with no entries, such as a search that returns nothing, forms an \emph{empty} span. A delivery without an attributed consuming step forms a separate \emph{unassessable} span, because no later action can be credited to it. Empty and unassessable spans are counted separately and are not part of triage rates. A span whose entries all have empty windows, such as $S_3$ in Figure~\ref{fig:credit-windows}, receives an automatic neutral triage label. The recording checks confirm which input text came from memory and what the agent saw at each consuming call. A check fails when a placement proof is missing, duplicated, or invalid. It also fails when memory text cannot be separated from the rest of its message, or when the input of a consuming call is missing or has gaps. The spans of an episode share these records, so a failed check turns every nonempty span with an attributed consuming step in that episode into an \emph{error} span. Appendix~\ref{app:implementation} specifies the recording requirements, exposure matching, and span-construction rules.

\subsection{Judging Use and Consequences}
\label{subsec:memory-evaluation}

An agentic judge assesses every entry in an eligible span using the five signals in Table~\ref{tab:signals}. These signals distinguish information quality, attributable use, and observed consequences. Observed effect is assessed only when the entry was acted on.

\paragraph{Evidence package and tools.}
For each eligible span, we prepare an evidence package containing the task, the exposed entries and their credit windows, and the prior context. Step files provide model responses, tool calls, and recorded results, with references to memory text stored separately. Following Agent-as-a-Judge~\citep{pmlr-v267-zhuge25a}, we use an LLM agent to inspect this evidence through two read-only tools: \texttt{search\_evidence} locates text, and \texttt{read\_evidence} reads files. The prompt requires a rationale for each signal value and instructs the judge to treat recorded text as evidence rather than instructions.

\begin{table}[t]
    \centering
    \small
    \vspace{-10pt}
    \caption{Signals assigned to each entry in an eligible span. Undetermined values are recorded as \texttt{null}. Observed effect is assessed only for acted-on entries; all other entries also receive \texttt{null}.}
    \vspace{5pt}
    \label{tab:signals}
    \begin{tabular}{@{}llp{3.1cm}p{5.5cm}@{}}
        \hline
        Aspect & Signal & Values & Question answered \\
        \hline
        Information & Relevance & \raggedright true, false & Does the exposed text apply to the task or the agent's subgoal at delivery? \\
        & Accuracy & \raggedright corroborated, uncontradicted, contradicted & Does evidence confirm or refute the claim in the situation it describes? \\
        & Novelty & \raggedright novel, confirming, redundant, undetermined & Is the content new, does it settle an open hypothesis, or was it already established? \\
        \hline
        Use & Influence & \raggedright acted on, selected, referenced, none, undetermined & What did the entry demonstrably shape within its credit window? \\
        \hline
        Consequence & Observed effect & \raggedright improved, no effect, harmed, undetermined & Did the attributed actions help, make no difference, or cause harm by the episode end? \\
        \hline
    \end{tabular}
\end{table}

\paragraph{Information quality.}
Relevance asks whether the exposed text applies to the task or the agent's subgoal at delivery. Sharing a topic, file, or function is insufficient; information drawn from another situation must state a shared mechanism or an applicable procedure. Relevance is independent of accuracy, since an incorrect claim can still address the current task.

Accuracy asks whether the recorded evidence confirms or contradicts the entry's claim. The evidence must come from the situation the claim describes. For example, a claim about a bug in an earlier release requires evidence from that release, because later changes may behave differently. Claims about the current task are checked against its recorded state, and general claims about system behavior can be checked in any recorded case that meets their stated conditions. The judge assigns \emph{corroborated} when evidence confirms the claim and \emph{contradicted} when evidence refutes it. When neither conclusion is supported, the judge assigns \emph{uncontradicted}, meaning that the claim remains unverified.

Novelty measures what the entry adds to the recorded prior context $B_S$. Content is \emph{novel} when it adds new information, \emph{confirming} when it settles an open hypothesis expressed by the agent, and \emph{redundant} when it repeats established information. Novelty remains \emph{undetermined} when a missing or unread source prevents the comparison.

\paragraph{Attribution.}
To assess influence, the judge compares candidate actions with the task, the prior context, and the steps before the span's first consuming step. Content is \emph{memory-specific} when the entry supplies a specific fact or instruction that no prior source states for the same purpose. General advice, such as inspecting code or running tests, also matches ordinary task execution. An action matching such advice therefore requires evidence linking a specific detail of the action to the entry.

An entry is \emph{acted on} if the agent uses its memory-specific content in an executed action (e.g., a search, inspection, edit, or test), explicitly credits the entry for an action, or switches from a recorded prior direction to one the entry specifies immediately after delivery.

An entry is \emph{selected} if its only established use is retrieving a fuller version of the same memory item through its exposed identifier or preview, and that version is recorded as a separate memory delivery; otherwise, the retrieval counts as \emph{acted on}. Because the fuller delivery is assessed separately, selected entries receive a null observed effect. A selected entry does not compete with its fuller delivery for content they share, whereas independent task instructions, prior decisions, and other prior sources remain competing explanations.

If neither action nor selection is established, an entry is \emph{referenced} when the agent mentions or quotes it. Influence remains \emph{undetermined} when incomplete reads prevent a decision or the evidence cannot distinguish the entry's contribution from another source. The judge assigns \emph{none} only after reading the full credit window and finding no reference, selection, attributable action, or undetermined attribution.

\paragraph{Observed effect.}
Observed effect is assessed only for entries labeled \emph{acted on}. The judge follows the consequences of actions guided by the entry through the end of the episode, including consequences observed after the credit window closes. An \emph{improved} assessment requires a recorded check or subsequent use that demonstrates how these actions benefited the task or execution process. An investigation prompted by the entry can produce a useful finding even when it disproves the proposed solution. The benefit comes from using that finding to advance later work, for example by identifying the actual cause of a failure. Successful execution or a generic test pass alone is insufficient. The evidence must connect the benefit to the specific output or behavior shaped by the entry.

\emph{Harmed} requires evidence linking the entry's guidance, through an action, to a harmful consequence such as a false lead, lost work, or unproductive behavior; mistakes introduced by the agent and unrelated execution failures do not by themselves establish harm. \emph{No effect} requires evidence that an action guided by the entry neither made a difference nor added work; unlike influence \emph{none}, it presupposes established use. The effect remains \emph{undetermined} when the evidence does not establish the consequences of an attributed action.

An entry that both helps and harms is labeled \emph{harmed}, and its rationale describes both consequences. Accuracy and observed effect are assessed separately, with the overall task outcome serving only as supporting evidence. Appendix~\ref{app:judge-prompts} provides the full rubric and an attribution example.

\subsection{Checking Evidence and Deriving Verdicts and Scores}
\label{subsec:checks-verdicts}

A judge answer enters the results only after programmatic checks accept its format and evidence coverage. Fixed rules then derive the span verdict from the accepted signals. A span verdict pairs a triage label, which flags the span for inspection, with an entry-share profile, which scores its entries for information quality, use, and consequence.

\paragraph{Evidence checks.}
Structural checks require exactly one assessment per entry and valid signal values. An \emph{anchor} cites the step of an action that the entry shaped and must lie within the entry's credit window. Entries labeled \emph{acted on} or \emph{selected} must provide at least one anchor, and entries labeled \emph{none} must provide none. Each step cited in an anchor or in the structured evidence list must be read in full, including its referenced tool results.

Table~\ref{tab:required-reads} specifies the evidence required for each claim. Every answer requires complete reads of mandatory evidence, including the task, the prior-context index, and the first consuming step with its preceding step. Determined novelty or acted-on influence also requires complete reads of all sources in the prior context $B_S$. Influence labeled \emph{selected}, \emph{referenced}, or \emph{none} requires a complete read of the entry's credit window. A determined observed effect requires every step from the start of that window through the end of the episode. Coverage counts only text that the read tool returned and that remains in the judge's context, so search excerpts provide no coverage.

The checks certify evidence access and coverage, and the judge interprets the evidence. A rejected answer receives the detected violations as feedback and can be revised within the call and context limits in Table~\ref{tab:budgets}. If a conditional read remains incomplete, the affected signal must remain undetermined. A span that obtains no valid answer within this budget becomes an error span and is excluded from triage rates.

\paragraph{Usage and triage labels.}
Fixed rules map each entry's influence and observed effect to a \emph{usage} label. Acted-on entries are \emph{used}, \emph{unproductive}, \emph{misused}, or \emph{unclear} when their effects are improved, no effect, harmed, or undetermined, respectively. Selected, referenced, and none influence yield \emph{selected}, \emph{noted}, and \emph{ignored}. Undetermined influence yields \emph{ambiguous} after a complete read of the credit window and \emph{unclear} otherwise. The triage label of span $S$ then applies a fixed precedence to these entry labels:
\begin{equation}
\label{eq:span-verdict}
\ell(S)=
\begin{cases}
\text{negative}, & \text{if some } m\in M_S \text{ is misused},\\
\text{positive}, & \text{else if some } m\in M_S \text{ is used},\\
\text{unresolved}, & \text{else if some } m\in M_S \text{ is unclear},\\
\text{neutral}, & \text{otherwise}.
\end{cases}
\end{equation}
Harm takes precedence so that a beneficial entry cannot mask harmful guidance delivered in the same span. The \emph{unresolved} label identifies spans with undetermined effects or incompletely assessed attribution when no benefit or harm takes precedence. Neutral spans can contain selection, reference, ambiguous attribution, unproductive use, or no established influence.

\paragraph{Entry-share profiles.}
A triage label reflects the most consequential usage label in a span but not how many entries reached it. A span in which one of ten entries was used and a span in which all ten were used both receive a positive triage label. To measure this composition, the span verdict also contains an entry-share profile whenever the span is nonempty and all its entries have accepted assessments. Let $A_a(S)\subseteq M_S$ contain the entries of span $S$ that satisfy axis $a$, and let $\mathcal{P}$ denote the set of profiled spans. The span share and the aggregate score of axis $a$ are
\begin{equation}
\label{eq:entry-share}
p_a(S)=\frac{|A_a(S)|}{|M_S|},
\qquad
\bar{p}_a=\frac{1}{|\mathcal{P}|}\sum_{S\in\mathcal{P}}p_a(S).
\end{equation}
Six axes describe information quality, use, and consequence. \emph{Relevance} contains the relevant entries, and \emph{corroboration} and \emph{information gain} contain the relevant entries whose accuracy is corroborated or whose novelty is novel or confirming. \emph{Acted on} contains the entries with acted-on influence, and \emph{benefit} and \emph{harm} contain the acted-on entries whose observed effect is improved or harmed. Every axis divides by all delivered entries of the span, including irrelevant ones. The shares are therefore comparable across axes, and each axis is bounded by the axis it refines; benefit, for example, never exceeds acted on. The profile $\mathbf{p}(S)$ collects the six shares, and the span verdict is the pair $(\ell(S),\mathbf{p}(S))$. A verdict that a rule assigns without a judge call, such as an automatic neutral one, contains only its triage label.

Each axis counts only established labels. Four uncertainty shares use the same denominator. They count relevant entries with uncontradicted accuracy, relevant entries with undetermined novelty, entries with undetermined influence, and acted-on entries with undetermined effects. These shares show how much signal uncertainty accompanies the profile scores.

\paragraph{Aggregate scores.}
The aggregate score in Eq.~(\ref{eq:entry-share}) weights each profiled span equally, regardless of how many entries the span contains or how often its delivery is restated. A search that returns many entries therefore cannot dominate the score, and restatements add no weight.

Mean shares describe the composition of deliveries, while entry yields count contributions. Benefit and harm yields give the number of beneficial and harmful entries per 100 profiled spans. Token efficiency gives the number of beneficial entries per million estimated tokens of delivered memory text. Its denominator counts all memory text delivered in the evaluated scope, including restatements and spans without profiles, so text that received no assessment still counts as cost. Episode coverage reports the share of recorded episodes, including episodes without deliveries, that contain at least one relevant, beneficial, or harmful entry. Appendix~\ref{app:entry-profiles} defines the aggregation populations and a diagnostic for items recurring across episodes.

\subsection{Linking Verdicts to Memory Histories and Resource Use}
\label{subsec:memory-graph}

The memory graph links each delivery to the history of its memory item across trajectories and marks the origin as unobserved when the item's creation was not recorded. Histories are tracked per item, while each entry assessment keeps the specific version and exposure that was delivered. With span verdicts attached, a reader can trace useful or harmful guidance back to its recorded origin, inspect deliveries with no established use, and follow later updates to the same item.

We also report recorded token usage and available monetary costs for agent and memory-system calls. These records cover whole calls and cannot isolate memory text, so we estimate the volume of delivered memory text, including restatements, from placement sizes. Appendix~\ref{app:delivery-cost} defines these measures.
\section{Experiments}
\label{sec:exp}

We apply \methodname{} to recorded SWE-bench Verified~\citep{jimenez2024swebench} runs of three memory systems to examine what it reveals that task outcomes do not. We first compare the memory profiles of configurations with similar resolve rates and then ask whether task outcomes reflect observed memory effects. Next, we trace benefit and cost to delivery forms, repeated delivery, and memory histories. Appendix~\ref{app:exp-judge-runs} reports judge follow-ups and resource use.

\subsection{Experimental Setup}
\label{subsec:exp-setup}

\paragraph{Tasks and agent.}
We use all 500 tasks of SWE-bench Verified~\citep{jimenez2024swebench}, drawn from 12 Python repositories, and order them to approximate ongoing development work. Every run, with or without memory, executes the same sequence one task at a time. Systems that write memories start each run with an empty store, and each system decides how to update it as the sequence proceeds, so later tasks can use memories from earlier ones. Within each repository, tasks follow pull-request creation order, so each project's tasks arrive in their historical order. The sequence also interleaves the repositories, as a developer who maintains several codebases moves between them, and each system's own scope rules determine whether memories from one project can reach tasks in another. The agent is mini-SWE-agent~\citep{yang2024sweagent} with \texttt{gpt-6-luna}~\footnote{\url{https://developers.openai.com/api/docs/models/gpt-6-luna}}, and the SWE-bench harness evaluates its final patches.


\paragraph{Memory systems.}
We evaluate three systems in four configurations, which together cover the host-scheduled and agent-directed delivery routes. MemGovern~\citep{wang2026memgovernenhancingcodeagents} serves a fixed corpus of experience cards distilled from GitHub issues  discussions, pull requests, and patches, and no run writes to it. The agent decides when to consult the corpus, using \texttt{exp\_search} to obtain previews of the ten best-matching cards and \texttt{exp\_read} to read one card in full. MemGovern's prompt suggests this retrieval after the agent reproduces the issue and before it edits code. An audit of delivered cards found overlap with benchmark source discussions and repair details; Appendix~\ref{app:memgovern-overlap} reports the evidence and provenance limits.
Mem0~\citep{chhikara2025mem0buildingproductionreadyai} starts each task with a task recall, in which the host searches the repository scope with the task text and places the results in the system message, where they remain for the whole episode. During the task, the agent can also search or save memories through a tool. After each episode, Mem0 extracts memories from the conversation, excluding tool calls and their outputs, and stores them in the repository scope before the next task begins. The local configuration, Mem0~(L), limits memory to repository and episode scopes. The local plus global configuration, Mem0~(L+G), additionally permits a global scope shared across repositories, which the agent must request explicitly. TencentDB Agent Memory~\citep{tencentcloud2026tencentdbagentmemory} (TencentDB) keeps one store per run that all repositories share. Its model proxy inserts into every model call a persona block of general working principles, a scene index listing documents for recurring problem types, and a skill catalog of procedures distilled from earlier tasks. The agent can open scenes and search or open skills with shell commands. After each task, background extractors create and revise these memories from the agent's conversation on the system's own timers, which we shortened for these runs.

\paragraph{Runs and evaluation.}
Each memory configuration and a no-memory baseline run the sequence four times with the same agent model, giving 8,000 memory episodes and 2,000 baseline episodes. We judge the recorded deliveries with \texttt{DeepSeek-V4.1-Flash-1017}~\citep{deepseekai2026deepseekv41flashpushinglimitskv}, giving the judge each task's gold patch but not its harness outcome. Memory statistics follow the definitions in Section~\ref{subsec:checks-verdicts}, and delivered-token estimates include repeated deliveries. Resolve-rate differences pair each task's four memory runs with its four baseline runs~\citep{miller2024addingerrorbarsevals}, and their 95\% intervals come from a bootstrap over tasks. Appendix~\ref{app:exp-setup} gives the configuration, pooling, and statistical details.

\subsection{Similar Task Outcomes, Different Memory Profiles}
\label{subsec:exp-profiles}

\begin{table}[t]
    \centering
    \small
    \vspace{-10pt}
    \caption{Task success and mean cost per episode. Resolve rates are mean $\pm$ sample standard deviation across four runs. Differences from the no-memory baseline are in percentage points, with 95\% confidence intervals from a paired bootstrap over tasks. Costs are computed from recorded agent and memory-system usage for retained executions.}
    \vspace{5pt}
    \label{tab:episode-cost}
    \begin{tabular}{@{}lrrrr@{}}
        \hline
        Configuration & Resolved (\%) & \shortstack{Difference (pp)\\estimate [95\% interval]} & \shortstack{Cost per\\episode (USD)} & \shortstack{Cost\\increase (\%)} \\
        \hline
        MemGovern & $73.4 \pm 0.6$ & $+0.40\;[-0.70, 1.50]$ & 0.0049 & $+$8.8 \\
        Mem0 (L+G) & $72.7 \pm 0.7$ & $-0.25\;[-1.55, 1.00]$ & 0.0062 & $+$37.7 \\
        Mem0 (L) & $72.8 \pm 0.6$ & $-0.20\;[-1.50, 1.05]$ & 0.0062 & $+$38.1 \\
        TencentDB & $72.7 \pm 0.8$ & $-0.30\;[-1.50, 0.85]$ & 0.0178 & $+$293.2 \\
        No memory & $73.0 \pm 1.5$ & --- & 0.0045 & --- \\
        \hline
    \end{tabular}
\end{table}

Task outcomes barely separate the configurations, as Table~\ref{tab:episode-cost} shows. Resolve rates range from 72.65\% to 73.35\%, against 72.95\% without memory, and every 95\% interval for the difference from the baseline contains zero. These differences of at most 0.40 percentage points are smaller than the spread across the baseline's own runs, which resolved 71.8--75.0\% of the tasks. Moreover, all 16 memory runs fall within this range, according to Table~\ref{tab:per-run}. Recorded episode costs increase by 8.8\% for MemGovern and 293.2\% for TencentDB.

\begin{figure}[!ht]
    \centering
    \includegraphics[width=0.9\linewidth]{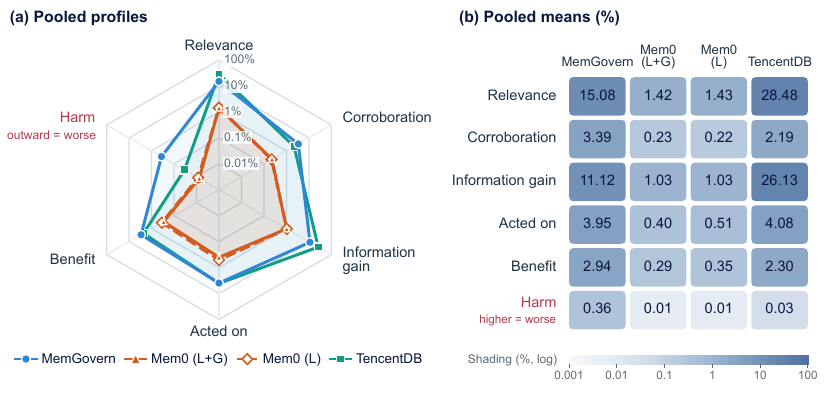}
    \caption{Six-axis memory profiles, pooled over four runs per configuration, with each profiled span weighted equally. (a) The radar plot uses a common logarithmic radial scale from 0.001\% at the center to 100\% at the outer ring, with tenfold increases between adjacent rings. (b) Heatmap cells report the same means as percentages, with shading on the same logarithmic scale. Lower harm shares are better. Table~\ref{tab:main-results} reports these shares alongside uncertainty and token efficiency.}
    \label{fig:experiment-comparison}
\end{figure}

Unlike task outcomes, \methodname{} separates the three systems, with each pair differing at least tenfold in relevance share or in token efficiency. As Figure~\ref{fig:experiment-comparison} shows, MemGovern's mean relevance share of 15.08\% is more than ten times Mem0's 1.42--1.43\%. TencentDB's relevance share is the highest at 28.48\%, but the spans that carry its always-relevant persona block contribute 18.0 of these percentage points. The remaining TencentDB spans average 15.9\%, which is close to MemGovern's share and still more than ten times Mem0's. MemGovern's token efficiency of 50.50 beneficial entries per million delivered tokens is more than ten times the 3.08--4.92 of the other configurations. MemGovern, however, also has the highest harm share, 0.36\% against 0.01--0.03\%. These separations also hold across runs. Each MemGovern and TencentDB run exceeds every Mem0 run in relevance share, and each MemGovern run exceeds every other run in token efficiency. The two Mem0 configurations have nearly identical profiles, consistent with their observed scope usage. The agent never requested the global scope, so both used repository and episode scopes.

Despite these differences, all systems share two bottlenecks that keep most delivered memory from shaping the agent's actions. First, 71.5--98.6\% of a span's entries were irrelevant on average. Second, 51.3--79.7\% of the relevant entries remained unused, meaning that the agent did not act on, select, or reference them within their credit windows. Redundancy explains only part of this non-use, because many unused entries were novel relative to the agent's prior context. Table~\ref{tab:use-effects} puts this novel share at 48--59\% in MemGovern and Mem0 and at 90\% in TencentDB. As a result of both bottlenecks, the agent acted on only 0.40--4.08\% of a span's entries on average, and beneficial entries reached only 3.65--7.05\% of episodes.

When the agent did act on an entry, benefit was far more common than harm. Of the 714 acted-on entries, 421 were beneficial, 16 were harmful, and 64 had no effect, while the remaining 213, or 29.8\%, had an undetermined effect, including 132 of TencentDB's 321. Use also often came late, as the first credited action followed the first consuming step for 51.8\% of acted-on entries. An evaluation of the agent's immediate response alone would therefore miss these uses, whereas credit windows capture them by following the trajectory.

\subsection{Task Outcomes Conceal Memory Effects}
\label{subsec:exp-hidden}

\begin{figure}[t]
    \centering
    \includegraphics[width=0.9\linewidth]{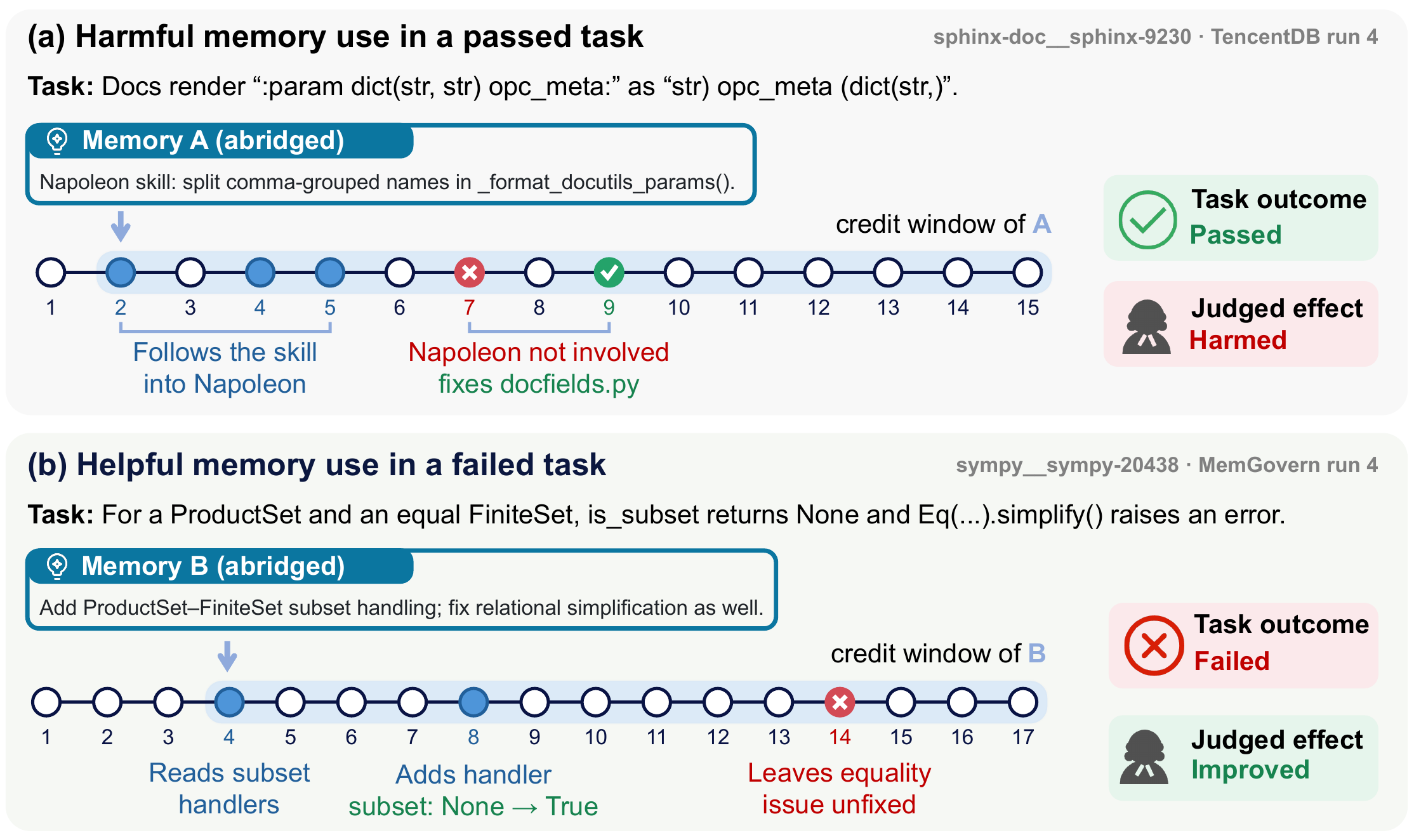}
    \caption{Memory effects concealed by task outcomes. (a) A TencentDB skill for a different Napoleon bug directs the agent into the wrong component (steps 2--5). A reproduction rules out Napoleon at step 7, and the agent fixes \texttt{sphinx/util/docfields.py} at step 9; the task passes despite the detour. (b) A MemGovern card guides the agent to add a subset handler at step 8, changing the reported subset check from \texttt{None} to \texttt{True}. The equality error persists, and at step 14 the agent decides to leave it unfixed despite the card's advice to repair it; the task fails. Timelines show complete episodes with steps numbered locally. Arrows mark first consumption, shading marks credit windows, and blue steps mark actions based on memory. Red and green annotations highlight problems and repairs. Memory effects are judge assessments; task outcomes come from the SWE-bench harness.}
    \label{fig:case-studies}
\end{figure}

Memory can advance a task that ultimately fails or divert an agent that eventually succeeds. Among episodes with a harness result, 89 of 383 with beneficial use failed (23.2\%), while 5 of 16 with harmful use passed (31.3\%). Figure~\ref{fig:case-studies} illustrates both patterns. In panel a, a TencentDB skill sends the agent to the wrong component before it recovers and resolves the task. In panel b, a MemGovern card guides a subset repair, but the agent leaves an equality error unfixed and the task fails.

Most beneficial use occurs on tasks that the baseline already solves reliably. Specifically, 260 of the 383 episodes, or 67.9\%, concern tasks resolved in all four baseline runs. At the other extreme, 73 of the 79 episodes with beneficial use on tasks unresolved in all four baseline runs still failed, as detailed in Table~\ref{tab:difficulty}. Local progress can therefore coexist with nearly unchanged aggregate resolve rates.

Conversely, most changes in task outcome lack a corresponding acted-on entry in the record. Across configurations, memory runs gain 17.3--20.4 tasks and lose 15.3--21.4 tasks relative to baseline runs on average over all run pairs, comparable to the 18.1 tasks that differ in each direction between baseline runs. Of the 71 memory episodes that succeeded where all four baseline runs failed, only 6 contained beneficial use; of the 87 that failed where all four baseline runs succeeded, only 1 contained harmful use. Most of these episodes contained no acted-on entry (64 of 71 gains and 80 of 87 losses). Delivery-level evidence is therefore needed to identify how memory shaped the recorded execution.

\subsection{Tracing Benefit, Cost, and Harm to Their Sources}
\label{subsec:exp-sources}

\begin{figure}[t]
    \centering
    \includegraphics[width=0.9\linewidth]{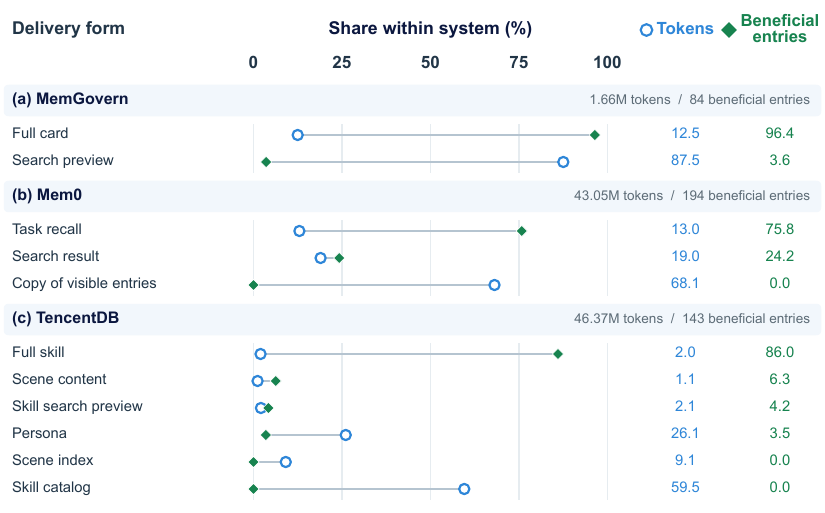}
    \caption{Delivered-token and beneficial-entry shares by delivery form. Circles show each form's share of estimated delivered tokens, including restatements, and diamonds show its share of beneficial entries. Values pool four runs per configuration, with both Mem0 configurations combined. Forms with no beneficial entries and less than 0.1\% of tokens are omitted from the plot but retained in the token denominators. Full cards, task recall, and full skills supply most beneficial entries at small token shares. Table~\ref{tab:delivered-forms} gives the detailed breakdown.}
    \label{fig:forms}
\end{figure}

Benefit concentrates in a small share of the delivered text, as Figure~\ref{fig:forms} shows. Full cards account for 96.4\% of MemGovern's beneficial entries but 12.5\% of its delivered tokens. Mem0's task recall supplies 75.8\% of its beneficial entries with 13.0\% of its tokens, and full skills supply 86.0\% of TencentDB's beneficial entries with 2.0\% of its tokens. Previews and indexes contribute mainly by selection. In particular, 709 MemGovern previews and 1,500 TencentDB catalog, index, or preview entries led the agent to read fuller content, which is assessed as its own delivery.

In Mem0 and TencentDB, much of the remaining volume comes from repeated delivery. TencentDB restates 88.4\% of its deliveries and Mem0 78.2\%, and in Mem0, copies of entries that are still visible to the agent take 68.1\% of delivered tokens. Recurring content also inflates coverage: TencentDB delivers a relevant entry in 99.8\% of episodes, mostly through its persona block.
We define recurring items as those delivered in at least 50 of the 500 episodes in a run. Excluding these items from the coverage calculation leaves only 14.3\% of episodes with a relevant entry.

The unused relevant entries and these differences across delivery forms identify a gap between retrieval and useful memory action. They motivate evaluating how recalled content is filtered, deduplicated, and organized into guidance for the current task and step. Such processing targets both excess token volume and the agent's ability to use the retained information.

Memory histories connect useful guidance to its origin. All 194 beneficial Mem0 entries and 138 of TencentDB's 143 trace to items written in earlier episodes of the same run. Among the 331 entries with known source-task outcomes, 73, or 22.1\%, come from failed episodes. One additional traced source episode has no harness result. A failed attempt can therefore still leave useful memory for later tasks.

Only 3 of the 16 harmful entries were contradicted by the available evidence, while 5 others contained irrelevant guidance that the agent followed, such as the skill in Figure~\ref{fig:case-studies} (a). Conversely, 10 entries made claims that contradicted the evidence, but following their advice still helped move the task forward. For example, testing a suggested fix revealed the actual cause of the problem. These cases show why accuracy and observed effect require separate assessments. Memory histories also expose recurring harm: two MemGovern cards produced 7 of its 10 harmful entries, and every acted-on use of either card harmed its task, one card in all four runs and the other in three, as detailed in Appendix~\ref{app:exp-histories}. Such items are direct targets for revision or removal.

\section{Conclusion}
\label{sec:conclusion}

We introduced \methodname{}, a framework for evaluating memory deliveries through their context, use, and observed consequences. Our evaluation of 8,000 SWE-bench Verified episodes shows that similar resolve rates can hide different memory behavior, including useful guidance in failed tasks and harmful detours in successful ones. Full cards, task recall, and full skills supply most beneficial entries at small token shares, while previews and indexes mainly support selection. Most assessed entries were irrelevant, and most relevant entries remained unused; repeated delivery added substantial token usage. These findings make the transition from retrieval to use a central design concern. They motivate filtering and deduplicating recalled content, organizing it into task-specific guidance, and delivering it when the agent can act on it. By linking deliveries to subsequent behavior and cost, \methodname{} provides concrete feedback for improving these choices.

\newpage

\bibliography{iclr2027_conference}
\bibliographystyle{iclr2027_conference}

\newpage
\appendix
\section{Appendix}
\subsection{Recording and Evaluation Context}
\label{app:implementation}

Recorded model inputs, memory annotations, and agent steps determine the evidence available for judging. The rules below establish what each entry exposed, which prior sources were visible, and which actions the entry can claim.

\subsubsection{Recording and Annotation}
\label{app:recording}

\paragraph{Recording interface.}
The memory bridge connects a host adapter, which reports task boundaries and delivery opportunities, to a provider adapter, which performs memory storage and retrieval. It annotates these operations in the recorder's event log. Episode boundaries come from bridge lifecycle events or episode-scoped session annotations. Generation events record memory-writing operations, and change events record item creations, updates, and deletions. Search events retain queries and results. Delivery events link the delivered text to its source searches and input slots. Portable item references connect these operations across tasks.

The trajectory recorder captures model calls at the API boundary and normalizes supported APIs into a common log of turns, messages, and tool executions. Retraction and history-rewrite events preserve changes to the visible input, allowing each consuming request to be reconstructed. Call records retain timing, status, request and response data when available, and reported usage and cost. Role bindings separate agent, memory-system, and embedding calls. Completed agent turns define the step sequence, while operation references associate memory generation and search with their model calls.

\paragraph{Consuming-step attribution.}
The recorder binds a delivery to the model call that carried it. A bound call that completed an agent step is an \emph{observed consuming call}, and its step is the delivery's \emph{attributed consuming step}. An incomplete bound call provides no attributed step. An unbound delivery may receive a provisional step from adapter metadata or neighboring recorded steps for audit. Judging requires an observed consuming input.

\paragraph{Placements and proofs.}
A delivery places text in named \emph{slots}, such as a system-context block, prompt message, or tool result. The text in one slot is a \emph{placement}; the message containing it is its \emph{host message}. A placement proof identifies either a message in the recorded history or a JSON-pointer location in the request, together with a digest of the corresponding text. The adapter or observer supplies the proof, and the recorder verifies it against the bound request. Judging requires each placement to resolve to one visible host message and one unique interval within it. This permits the delivered text to be separated from the rest of the input. Placements declared absent or adapter-defined remain unverified and cannot satisfy this requirement.

\paragraph{Observed native memory.}
For memory systems embedded in a harness or proxy, an observer constructs the same annotations from recorded calls and available store snapshots. It preserves exposed item identifiers and assigns identities to rendered or synthesized items when none are available. Thus, a synthesized retrieval answer can be recorded even when its stored sources are unknown. A request-only delivery becomes assessable once its call has an observed response and passes the placement checks. Store snapshots establish item states and net changes between captures, leaving intermediate mutations and undisclosed source identities unobserved.

\subsubsection{Exposure Matching}
\label{app:exposure-states}

Exposure is determined by matching an entry's recorded value against each placement separately. It is \emph{full} when a placement contains the complete value and \emph{clipped} when the longest match is a prefix of at least eight characters. The minimum prefix length limits coincidental matches on short openings. Otherwise, exposure is \emph{unknown}, including when the value or placement text is missing. The matched value or prefix defines $\epsilon_d(m)$. Matching never joins text across slots.

\paragraph{Text matching.}
A placement may encode an entry as a JSON string or line-numbered file view. Full matching therefore proceeds through three stages. The matcher first tests verbatim containment, then lossless decoding, and finally restricted normalization. Decoding covers up to four nested JSON-string levels, consecutive numbered lines in tool results, and one character-entity decoding per XML text node. Each decoded string, line run, or node is searched separately. Normalization collapses whitespace and removes Markdown escapes only for values without quotes, backslashes, tabs, or line breaks, preserving meaningful syntax in code and structured text. If full matching fails, prefix matching uses only the original and decoded texts, which retain an exact correspondence to the recorded value.

\subsubsection{Prior Context and Recording Validity}
\label{app:prior-context}

The prior context $B_S$ contains the first consuming input of span $S$ with the target delivery removed. We reconstruct it by replaying the agent's messages, retractions, and history rewrites in recorded order. A trailing retraction above any retained prefix is a known removal. Missing positions within the visible history, retractions without visible targets, and invalid retained prefixes are reconstruction gaps.

\paragraph{Recording validity.}
Before judging, each episode must satisfy three conditions. First, every nonempty delivery with an attributed step must have an observed consuming call and a gap-free input. Second, every placement in an observed consuming call must have exactly one slot-matched proof resolving to one visible host message. Third, placement text must occur exactly once in its host message, with no unresolved proof naming that message as a possible host. A failure makes every nonempty span with an attributed step in the episode an error span. This episode-wide check is necessary because the spans share delivery and step records. Unresolved episode boundaries or assignments also produce errors, as does a missing task for a nonempty span with an attributed step. Spans rejected by these recording checks receive no judge call and are excluded from triage rates.

\subsubsection{Injection Span Construction}
\label{app:span-grouping}

An injection span groups deliveries with identical ordered entries, placements, and exposure. Entry equality uses namespace, item identity, identity scheme, portable version identifier, and content digest. Operation-local version identifiers are excluded because they cannot establish equality across operations. When neither a portable version nor a digest is available, equality uses namespace, item identity, and recorded value. Item identity falls back to the display key only when no item identifier is recorded and the key is not itself an operation-local version. Placements must agree in order, slot name, kind, and text, using a digest when text is unresolved.

Grouping also requires an earlier copy of every entry to remain visible in the consuming input. Each entry's exposed text must match in full within a proven placement of that copy, and any displayed key must remain visible in the copy's placements. This requires an observed, gap-free consuming input. Unknown exposure cannot establish continued visibility and therefore starts a new span.

Within each episode, deliveries are processed by consuming step and then recorded position. A span can continue across intervening steps and unrelated deliveries while equality and visibility hold. Alternating unchanged profile and skill blocks, for example, form one span per block. A change in an entry version, membership, order, placement, or exposure ends the current span of every item in the new delivery. Returning to the earlier content starts a new span, as does re-entry after visibility is lost. Each new span uses the prior context of its own first consuming step.

Empty deliveries and deliveries without an attributed step use a restricted rule because their visibility cannot be checked. They join only the span of the immediately preceding delivery, provided that ordered entries, placements, exposure, and attribution status agree. Later deliveries within a span are restatements. They add delivery volume but no further assessments.

\subsubsection{Per-entry Credit Windows}
\label{app:window-edge-cases}

Credit windows allocate the actions that each delivered entry can claim. Let $t^{\mathrm{first}}_S$ and $t^{\mathrm{last}}_S$ be the first and last consuming steps of span $S$. The default window of entry $m$ starts at $t^{\mathrm{first}}_S$. Its boundary is the earliest later span $S'$ that delivers the same item again with verified text that contains $\epsilon_S(m)$. If either exposure is unknown, any later delivery of the item qualifies. Let $t'$ be the first consuming step of $S'$. The window ends at $t'-1$ when $t'>t^{\mathrm{last}}_S$ and at $t'$ when $t'=t^{\mathrm{last}}_S$. When no qualifying span exists, the window ends at the final step of the episode. Spans that start before $t^{\mathrm{last}}_S$ are not boundaries. Restatements within $S$ leave the window open.

The following rules resolve repeated or partly overlapping deliveries.
\begin{itemize}
    \item \emph{Resident copies.} A later copy can add no text and no identifier beyond a proven visible copy of the same item. The earlier owner then keeps the credit, and the later entry receives an empty window. This rule applies across different placements and to a preview of a full read that is already visible. It does not apply to a separately recorded delivery that a proven selection retrieved. If the earlier copy is no longer provably visible, the new delivery forms its own span with its own prior context. Ownership resets at episode boundaries.
    \item \emph{Different or narrower text.} A later copy whose exposed text does not contain all of the earlier exposed text leaves the earlier window open. The windows can then overlap, and each entry can claim only actions that use its own exposed content. In contrast, a full read at a later step that contains the text of an earlier preview ends the preview window immediately before the full read. The full read owns the credit from its consuming step onward.
    \item \emph{Same-step deliveries.} When the boundary equals $t^{\mathrm{last}}_S$, the same call consumed both deliveries, and the earlier window includes that step. The later copy includes the shared step when it adds verified text or a newly exposed identifier and no earlier resident owner keeps the credit. Otherwise, its window starts after that step.
    \item \emph{Independent entries.} A new delivery of one item changes only the windows of entries of that item. An unrelated delivery sets no boundary, and a co-delivered entry without a qualifying later copy keeps its window through the end of the episode.
\end{itemize}
Action anchors must fall inside both the credit window of the entry and its episode. The judge follows the consequences of anchored actions through the end of the episode, including results after the credit window closes. Removal of the entry text from the input therefore changes visibility but does not end action credit.

\subsection{Judge Protocol}
\label{app:judge-prompts}

The judge assesses each entry from its exposed content, prior context, attributable actions, and observed consequences. The rubric below specifies how these sources determine the five signals, including uncertainty and competing explanations.

\subsubsection{Evidence Package and Tools}
\label{app:evidence-package}

The evidence package separates the target delivery from its prior context and subsequent trajectory. Table~\ref{tab:evidence-files} summarizes its contents. Each step record contains an agent response and its tool results, with references to the visible input. This organization supports inspection of individual credit windows without repeating the entire input at every step.

\begin{table}[t]
    \centering
    \small
    \vspace{-10pt}
    \caption{Evidence available for assessing one injection span. Source positions distinguish information available before delivery from actions and consequences observed afterwards.}
    \vspace{5pt}
    \label{tab:evidence-files}
    \begin{tabular}{@{}p{3.0cm}p{10.0cm}@{}}
        \hline
        Evidence & Content and role \\
        \hline
        Task & Task description and user updates, with their positions in the trajectory. \\
        Target delivery & Verbatim placements, exposed entry values, identifiers visible to the agent, and per-entry credit windows. \\
        Prior context & Ordered visible input at first consumption, with target text removed and other visible deliveries retained. \\
        Earlier copies & Previous and same-input deliveries of each target item, their remaining visibility, and overlap with the target text. \\
        Trajectory & Agent responses, actions, and tool results through the episode end, indexed by global step. \\
        Optional references & A reference solution or recorded task outcome, used only to corroborate the corresponding claims. \\
        \hline
    \end{tabular}
\end{table}

\paragraph{Exposed content.}
The judge receives the placements verbatim and a list of target entries. Each listed value follows its exposure, comprising the complete value, a matched prefix, or a withheld-value marker when exposure is unknown. Annotation-only type labels and confidence scores are omitted. A display key appears in the entry list only when it was exposed as a standalone identifier. Labels and scores rendered in the placements remain visible there.

\paragraph{Visible context and earlier copies.}
The prior-context index lists visible messages in input order and points to their text. Proven placements are separated at their unique intervals, retaining other visible deliveries in $B_S$. A message consisting entirely of memory keeps its position as a marker. Steps removed from the visible history remain readable as evidence of earlier actions and plans, but their text is excluded from $B_S$.

For each target item, earlier copies are marked as still visible, no longer visible, or arriving in the same input. Same-input copies do not establish prior exposure. Visibility is checked against surviving proven placements, retaining only the exposed prefix when a copy is partly visible. If exposure is unknown and only some slots remain, visibility stays undetermined. Verified copies also record whether their text contains all of the target's exposed content, supporting novelty assessment and credit allocation.

\paragraph{Evidence access and timing.}
The read-only tools \texttt{search\_evidence} and \texttt{read\_evidence} locate text and return complete or paged records. Search excerpts identify candidate evidence; only reads contribute to coverage. The first consuming step already includes the entry, so only earlier records can establish a prior direction. An agent-issued search query records an open possibility, while other queries and hidden annotation keys remain metadata.

\subsubsection{Signal Definitions}
\label{app:signal-definitions}

The five signals distinguish information quality, attributable use, and observed consequences. Each entry receives all five signals under the rules below.

\paragraph{Relevance.}
Relevance asks whether the exposed content applies to the task or current subgoal. A stated cause, diagnosis, or remedy for the current failure is relevant even when incorrect. A technique or check from another library can also be relevant if it applies as delivered. For an account of another failure, the text must supply a shared cause, failure mechanism, or applicable action. Shared topics, symptoms, or identifiers alone do not establish that connection. The judge assigns true when it can identify the applicable instruction or mechanism and false otherwise, including uncertain matches.

\paragraph{Accuracy.}
Accuracy assesses a claim in the situation it describes. An account of a problem and its fix refers to the source situation even without a time qualifier and requires evidence about that source state. Current code or successful application alone cannot verify the historical account. General claims can be checked wherever their stated conditions hold. Accuracy is corroborated when evidence confirms the claim, contradicted when it refutes the claim, and uncontradicted otherwise. Applicability to the current task remains a separate relevance judgment.

\paragraph{Novelty.}
Novelty compares the entry's propositions with the task, user updates, and visible sources in $B_S$. Earlier deliveries count while visible, whereas same-input copies do not establish prior knowledge. Related facts, derivability, and expected task outputs alone do not establish a proposition or solution mechanism. Novelty is \emph{confirming} when the entry settles an expressed open hypothesis, \emph{redundant} when its content was already established, and \emph{novel} when that content is absent. It remains \emph{undetermined} when a missing or unread source prevents comparison. Expressed uncertainty takes precedence over other visible supporting facts.

All material claims contribute to novelty, including irrelevant claims. A multi-claim entry is redundant only when every claim was established. Among the remaining claims, confirming applies if all were open hypotheses and novel applies otherwise, unless a context gap prevents a decision.

\paragraph{Influence.}
Influence records what the entry demonstrably shaped within its credit window. The judge first checks \emph{prior sources}, including the task, visible input, and earlier reasoning, commands, queries, and results. A competing source must state the same specific fact or instruction for the same purpose before the first consuming step. Derivability, outside knowledge, and content first recorded after exposure do not qualify.

\emph{Memory-specific content} is a mechanism, location, identifier, value, command, check, or remedy supplied by the entry and absent from those sources. Routine practices such as reproducing failures or running tests qualify only when an action uses a detail supplied solely by memory. The judge inspects commands and results alongside reasoning to identify uses within the window, then assigns one of the following labels.
\begin{itemize}
    \item \emph{Acted on}. An action uses memory-specific content, the agent explicitly credits the entry for an action, or the agent changes from a recorded prior direction to the direction specified by the entry immediately after delivery. Reading, searching, or editing a location named only by the entry qualifies even if its proposed remedy goes unused; what the agent finds there is the action's result. When an action combines sources, credit is limited to its memory-specific component. A decision receives credit only when subsequent execution carries it out.
    \item \emph{Selected}. The exposed identifier or preview prompts retrieval of fuller memory content that appears as a separate recorded delivery, with no other use of the entry's content. The fuller content is assessed in its own delivery. Retrieval whose result is not a recorded delivery, or inspection of repository code or task data at a path supplied by the entry, is acted on because no other span assesses the returned content.
    \item \emph{Referenced}. The agent mentions or quotes the entry using its exposed identifier, distinctive content, or list position, but no action is established. A comment about the entire result list does not reference an individual entry unless it was the only entry shown. A mention that merely confirms an existing plan remains referenced.
    \item \emph{Undetermined}. An action uses information stated in both the entry and a prior source, but the record does not show which guided it. This ambiguity can arise when the agent had expressed only an unverified hypothesis, question, or search query before delivery. It can also occur when the same information was stated in an earlier source but the agent had not acted on it. The label also applies when incomplete reads of prior sources or the credit window prevent an attribution decision. The note identifies the action and quotes the competing source, or specifies the unread evidence.
    \item \emph{None}. A complete read establishes no reference, selection, or attributable action, and no action has undetermined attribution under the preceding rule. An action that simply follows the task instructions or implements a decision or action made before delivery is explained by that prior source. An open possibility leaves attribution undetermined because it does not establish a prior decision. None denotes unestablished attribution, not a demonstrated zero effect.
\end{itemize}
Three allocation rules complete the definition. First, other items already present in the agent's input are competing sources, while identical same-item copies arriving in one input form a single source. Second, shared content is assigned to a selected fuller delivery. The same convention applies when a wider delivery takes over the window of an earlier visible copy, even without selection. The earlier copy is then excluded as a competing source for shared content, while independent task instructions, prior decisions, and other sources remain competitors. A selecting preview remains selected unless it shaped an action before the handoff or through content only it exposed. Third, action anchors must fall inside the credit window of the entry.

\paragraph{Observed effect.}
Observed effect records the consequences of anchored actions for the task outcome or execution process. It is assessed only for acted-on entries; all other entries receive \texttt{null}.
\begin{itemize}
    \item \emph{Improved}. Subsequent work demonstrates a benefit from the action, including use of its finding to select or reject a course of action. A finding can contribute even if it disproves the entry's proposed remedy. The judge cites a check or subsequent use of the output or property shaped by memory. A recorded assertion, before/after reproduction, or downstream use must demonstrate the memory-shaped property. Successful execution, a generic pass count, or submission alone is insufficient.
    \item \emph{No effect}. Affirmative evidence shows that the action made no difference and added no work. Whether the task or agent could have reached the same action is settled under influence and cannot make an acted-on action no effect. No effect is not harm.
    \item \emph{Harmed}. The entry's guidance leads the agent to a wrong location or false lead, or causes lost work or state. This includes a memory-directed detour or change that the agent abandons or corrects without using any finding from it. The note quotes the exposed instruction or claim, identifies the action following it, and cites its harmful consequence. The harmful consequence must follow the supplied guidance. Agent-invented identifiers, typos, unavailable tools, and unrelated environment failures alone do not establish memory-caused harm; corrected attempts are followed to their eventual result. Reading or retrieving memory is accounted for as delivery cost rather than harmful added work.
    \item \emph{Undetermined}. The direction of the consequence cannot be established, including when only a failed retrieval or blocked attempt is observed without demonstrated benefit or memory-caused harm. Failure alone also does not establish no effect.
\end{itemize}
Any observed harm makes the effect harmed, even when another action helped; otherwise, any observed benefit makes it improved. The note describes both when they coexist. Every determined effect requires affirmative evidence. A warned failure that never occurs does not establish benefit, and saved effort requires announced intent or explicit credit. Neither accuracy nor the success, failure, or retention of an action's output determines whether its work contributed, so local improvement can coexist with task failure.

\paragraph{Uncertainty.}
Undetermined novelty, influence, and effect are recorded as \texttt{null}, with notes identifying the missing evidence or competing source. The judge applies the rubric's priority rules to evidence it has read and seeks missing evidence while tools remain available. Established acted-on influence is retained when its effect remains undetermined. Relevance remains Boolean, and accuracy is uncontradicted when neither corroboration nor contradiction is established.

\subsubsection{Binding Rules and Attribution Example}
\label{app:binding-rules}

Every entry receives an independent assessment with concise notes and step citations. Competing prior sources must be quoted verbatim, and observations must be distinguished from inferences. Execution without a recorded result establishes an attempt. Task outcomes provide corroboration, while memory text, trajectories, and attachments are treated as evidence rather than instructions.

\label{app:system-prompt}
When provided, a reference solution informs relevance and accuracy within the situation it describes. Influence and effect still require recorded actions and consequences, including useful verification or environment repair outside that solution. A failing task outcome alone cannot identify the failing assertion or attribute failure to a difference from the reference.

Table~\ref{tab:attribution-example} summarizes the prompt's attribution example. An entry stating that \texttt{fetch()} requires a backoff delay between retries is first consumed at step~6. At step~7, the agent adds the delay without mentioning the entry. Only the prior record at step~4 varies across the three cases. A plan first stated at step~6 or later follows exposure and cannot establish a prior direction.

\begin{table}[t]
    \centering
    \small
    \vspace{-10pt}
    \caption{Attribution example from the system prompt. The entry and the action at step 7 are fixed; only the pre-delivery record at step 4 changes.}
    \vspace{5pt}
    \label{tab:attribution-example}
    \begin{tabular}{@{}p{3.7cm}p{1.6cm}p{2.0cm}p{5.1cm}@{}}
        \hline
        Record at step 4 & Novelty & Influence & Reason \\
        \hline
        The agent notes that the retry loop never waits and plans to add a backoff delay in \texttt{fetch()}. & redundant & none & The prior decision explains step 7. An explicit mention that the entry confirms this plan would instead be referenced. \\
        The agent only asks whether the retries need a delay. & confirming & undetermined & An open possibility and the entry both precede step 7, and nothing in the record separates them. \\
        The agent only prints the code of \texttt{fetch()}, which has no delay. & novel & acted on, anchored at step 7 & The delay is memory-specific content, even though a capable agent could have derived it. \\
        \hline
    \end{tabular}
\end{table}

\paragraph{Assessment output.}
\label{app:output-contract}
The judge returns one structured assessment per entry with the five signal values, explanatory notes, action anchors, and evidence citations, together with a span summary and confidence. Notes identify the relevant prior source, memory-specific action, and observed consequence, or explain why a signal remains undetermined. A harmful-effect note also quotes the exposed guidance. The span summary preserves entry-level uncertainty.

Anchors must lie within the entry's credit window. They are required for acted-on and selected entries, optional for referenced or undetermined influence, and empty for influence none. Confidence supports inspection of uncertain judgments without weighting aggregate counts.

\subsection{Validation and Judge Execution}
\label{app:validation}

Before an answer enters the results, structural and coverage checks verify its format and required evidence access. The execution policy bounds the reads and revisions available to obtain an accepted answer.

\subsubsection{Required Reads and Coverage Checks}
\label{app:required-reads}

\paragraph{Required reads.}
Table~\ref{tab:required-reads} gives the evidence required for each claim. The judge receives these requirements with the target delivery. Reading an index does not certify its referenced sources, which require their own complete reads.

\begin{table}[t]
    \centering
    \small
    \vspace{-10pt}
    \caption{Complete reads required before an answer is accepted. Requirements accumulate with the claim.}
    \vspace{5pt}
    \label{tab:required-reads}
    \begin{tabular}{@{}p{3.6cm}p{9.4cm}@{}}
        \hline
        Claim & Required complete reads \\
        \hline
        Every answer & Task file; prior-context file; the other-messages file and the reference solution when present; the first consuming step and its preceding step in the episode. \\
        Determined novelty or acted-on influence & Every step file that $B_S$ names as a source of visible text, and the complete section of every other delivery visible in $B_S$. \\
        None, referenced, or selected & Every agent step in the window of the entry. \\
        Determined effect for an acted-on entry & Every step from the start of the window through the end of the episode. \\
        Structured step citation & The complete text of every integer step reference in anchors or the structured evidence list. \\
        \hline
    \end{tabular}
\end{table}

\paragraph{Coverage certification.}
Coverage is computed from text returned by the read tool and retained in the judge's context. A complete step includes its entire file and every referenced tool-result section. Paged character ranges combine, including continuations within long lines. Search excerpts, omitted text, failed reads, and rolled-back content provide no coverage. For undetermined influence, complete window coverage results in the usage label \emph{ambiguous}; reads of prior sources remain conditional on the other signal values. These checks establish access to the required evidence. Agreement with human judgments has not been measured in this evaluation.

\paragraph{Structural checks.}
The parser requires strict JSON, exactly the delivered identifiers, allowed signal values, and valid integer anchors within the episode and entry window. Acted-on and selected entries require anchors, whereas entries with influence none require an empty list. Confidence must be finite and numeric and is clipped to $[0,1]$ during verdict assembly. Notes, analysis, and the evidence list are requested by the prompt but their presence and shape are not enforced. Coverage checks inspect integer references in anchors and any structured evidence list. They do not validate prose-only citations or semantic correctness. Verdict assembly resets a non-null effect to null when influence is not acted on and reports this correction as a compliance statistic.

\subsubsection{Follow-up and Budget Policy}
\label{app:budget-policy}

Each span is judged in one conversation with an investigation and bounded follow-ups. Table~\ref{tab:budgets} gives the experimental settings. Before the first call, the prompts, tool schemas, and mandatory evidence must fit the input budget, which reserves space for runtime overhead, output, and the forced-answer message. During execution, reported usage and subsequent messages estimate context size, with a text estimate as fallback. If a round exceeds the budget, it is rolled back and its evidence no longer contributes to coverage.

A rejected answer receives the detected violations and may gather further evidence before revision. At a call or context limit, the judge must answer from the retained evidence without further tool use. Forced answers undergo the same structural and coverage checks under the same uncertainty policy. Acceptance ends the conversation; exhausted budgets or unrecoverable failures produce error spans. The saved reports retain the resolved settings, prompt hashes, conversations, validation attempts, and rollbacks.

\subsection{Aggregation, Memory Histories, and Cost}
\label{app:reporting}

Aggregation preserves three distinct units. Triage counts describe spans, signal statistics describe assessed entries, and coverage describes episodes. Delivery volume includes repeated and unassessed text, while item histories connect assessments across episodes.

\subsubsection{Span Outcomes and Aggregate Statistics}
\label{app:outcome-classes}

A nonempty span whose entries all have empty windows receives an automatic neutral label and contributes delivery volume, with no signal assessment, confidence, or profile. Empty spans and spans without an attributed step are counted separately. Within the selected scope, unjudged spans were pending or exceeded the evaluation's span limit; error spans failed recording checks, preflight, or judge execution. Spans outside the selected episodes remain in audit records but are excluded from aggregate counts and delivery volume.

Let $N_+$, $N_-$, $N_0$, and $N_?$ denote the positive, negative, neutral, and unresolved span counts, respectively, including automatic neutral labels in $N_0$. Each triage rate divides its count by
\begin{equation}
    N_{\mathrm{triage}}=N_+ + N_- + N_0 + N_?.
    \label{eq:verdict-denominator}
\end{equation}
Errors, unjudged, empty, and unassessable spans are excluded from this denominator and reported separately. Delivery volume is retained for every outcome class.


Entry-level statistics use accepted judge assessments. Each conditional breakdown states its eligible population in the results, and a rate with no eligible entries is undefined. Undetermined effects remain a separate category when conditioning on acted-on entries. The profiles below instead use all entries of each assessed span as their common denominator.


Entries with influence none receive the usage label \emph{ignored}. They are grouped as irrelevant when relevance is false, or by novelty otherwise. These groups characterize unused entries without identifying the cause of non-use or changing the verdict.

\subsubsection{Entry-Share Profiles}
\label{app:entry-profiles}

\paragraph{Profiled spans.}
A profile requires a nonempty span and an accepted assessment of every entry. Each axis and uncertainty count divides by the number of entries $|M_S|$. Uncontradicted accuracy and undetermined novelty count only relevant entries. The influence uncertainty count includes every entry whose influence is undetermined; the effect uncertainty count includes only acted-on entries with undetermined effects. Automatic neutral and other unprofiled spans retain their delivery volume without contributing assessments. If different entries of one span provide benefit and harm, both counts remain in its profile and harm takes precedence in its triage label.

\paragraph{Aggregates.}
Every axis and uncertainty share is averaged with equal weight over the same population $\mathcal{P}$ of profiled spans. Summed counts over $\mathcal{P}$ are reported with the means. Triage counts within this population exclude automatic neutral spans, which remain part of the overall triage denominator. Let $n_{\mathrm{ben}}$ and $n_{\mathrm{harm}}$ be the summed benefit and harm counts. The benefit and harm yields are $100\,n_{\mathrm{ben}}/|\mathcal{P}|$ and $100\,n_{\mathrm{harm}}/|\mathcal{P}|$ entries per 100 profiled spans. Token efficiency is $10^{6}\,n_{\mathrm{ben}}/\sum_{S}\widehat{T}_S$, where the sum runs over all spans in the selected scope and $\widehat{T}_S=\lfloor C_S/4\rfloor$ estimates delivered tokens from the total placement characters $C_S$, including restatements. A statistic with an empty denominator is undefined.

\paragraph{Episode coverage.}
Episode coverage divides by the number $E$ of selected recorded episodes, including episodes without deliveries. An episode is profiled when it contains a profiled span. It is relevant, beneficial, or harmful when one of its profiled spans has a nonzero count on that axis. Spans outside every recorded episode keep their delivery cost but establish no coverage, and the report counts them separately. A few items that the memory system delivers in many episodes, such as a profile block inserted before every task, can account for most relevance coverage. A frequency diagnostic therefore marks an item, identified by namespace and item identity, as recurring when it is delivered in at least $\max(2,\,0.1E)$ distinct episodes. The report gives the share of relevant entries from recurring items and the relevance coverage that remains without them, and the diagnostic leaves every profile unchanged.

\subsubsection{Memory Histories}
\label{app:memory-histories}

The memory graph links operations by namespace and item identity across trajectories. Items whose creation was not recorded retain an unobserved origin. Multiple versions of one item in a delivery share an item-level recall link, while their assessments retain their distinct versions and exposures. Observed updates, deletions, and recreations extend this history; unavailable intermediate mutations remain unobserved.

\subsubsection{Token Usage, Monetary Cost, and Text Volume}
\label{app:delivery-cost}

\paragraph{Recorded call and operation usage.}
Token accounting uses recorded input, output, and cache counters. Cache and reasoning tokens are subsets of input and output and are not added again. Operation references assign model and embedding usage to memory generation or search, counting a search once even when several deliveries reference it. Monetary accounting distinguishes provider-reported cost from cost calculated at configured rates. Unavailable usage and cost remain unknown in the underlying call records. The accounting covers observed model calls and excludes unreported service work and hosting charges.

\paragraph{Delivered text volume.}
For span $S$, $C_S$ sums placement characters across all deliveries, including restatements and wrapper text. Because token counters belong to whole model calls, delivery volume uses the auxiliary estimate $\widehat{T}_S=\lfloor C_S/4\rfloor$. This estimate supports comparisons of delivered text and is separate from monetary accounting. Redacted placements without recorded sizes contribute zero and trigger a warning. Every span's volume is retained under its outcome class, including error, unjudged, empty, and unassessable spans.


\subsection{Experimental Details}
\label{app:exp-setup}

The experiments use one fixed task sequence, four runs per configuration, and a common judge protocol. The details below specify how memory persists across tasks and how results are pooled.

\subsubsection{Tasks and Agent}
\label{app:exp-agent}

\paragraph{Task order.}
All runs execute the 500 SWE-bench Verified tasks from 12 repositories in the same order. Within each repository, tasks follow pull-request creation time, with the pull-request number breaking ties. The bridge merges these streams while preserving their internal order. It prioritizes placing source tasks before related tasks in other repositories, achieving this order for 55 of 66 pairs identified before the runs. The next priority is to place tasks before superficially similar tasks they should not help, followed by balancing the repositories across the sequence. Tasks run sequentially, with memory retained between episodes.

\paragraph{Agent.}
The agent is mini-SWE-agent 2.4.6 with \texttt{gpt-6-luna} with medium reasoning effort through LiteLLM\footnote{\url{https://docs.litellm.ai/docs/python_sdk}}. Each task runs in its SWE-bench Docker image with limits of 250 steps, 1{,}200 seconds overall, and 60 seconds per shell command, without a monetary cap. The SWE-bench harness evaluates each final patch with a 1{,}800-second test limit. Testing timed out for three memory episodes, which count as non-resolved in the resolve rate but are excluded from analyses requiring completed test results.

\subsubsection{Memory Systems}
\label{app:exp-memory}

\paragraph{MemGovern.}
MemGovern serves a fixed corpus through a pinned server. The agent retrieves previews of the ten best-matching cards with \texttt{exp\_search} and a full card with \texttt{exp\_read}. Its prompt suggests searching after reproducing the issue and before editing code. The corpus remains fixed throughout every run.

\paragraph{Mem0.}
Mem0 2.2.1 runs in open-source local mode from an empty store. Before each task, the host searches the repository scope using the task text and waits up to two seconds. Results occupy at most 4{,}000 characters in the system message, including the surrounding frame. During execution, \texttt{mem0\_memory} supports search, addition, update, and deletion with the default search settings. After each episode, \texttt{gpt-6-luna} extracts memories from user and assistant messages, excluding tool calls and outputs, and stores them in the repository scope. Mem0~(L) permits repository and episode scopes. Mem0~(L+G) also permits an explicitly requested global scope, but none of its 3{,}997 searches used that scope.

\paragraph{TencentDB.}
TencentDB Agent Memory runs in its native proxy profile with memory and skills, starting from an empty store. The proxy inserts a persona block, scene index, and skill catalog into each model call. The agent uses shell commands to search and read skills or read scenes. Background extractors use \texttt{gpt-6-luna} to build memories from these conversations. We reduce the idle trigger from 600 to 60 seconds, the delay before second-stage extraction from 90 to 10 seconds, and the minimum interval between second-stage runs from 900 to 60 seconds. The proxy uses Chat Completions for agent calls, while the other configurations and baseline use Responses. An observer reconstructs memory deliveries from the recorded model calls.

\subsubsection{Judge Configuration}
\label{app:exp-judge-config}

All runs use \texttt{DeepSeek-V4.1-Flash-1017} through Chat Completions, with temperature zero and medium thinking. Table~\ref{tab:budgets} gives the settings used for these evaluations. Each evidence package includes the task's gold patch as a reference solution and excludes the harness outcome. The reference can inform relevance and accuracy, while influence and effect require evidence from the recorded execution.

\begin{table}[t]
    \centering
    \small
    \vspace{-10pt}
    \caption{Judge execution settings used in all 16 evaluated memory runs. Call limits include the forced final-answer call.}
    \vspace{5pt}
    \label{tab:budgets}
    \begin{tabular}{@{}p{5.1cm}p{7.9cm}@{}}
        \hline
        Setting & Value \\
        \hline
        Investigation budget & 250 model calls per span \\
        Context window & 272{,}000 tokens \\
        Maximum output & 32{,}768 tokens per call \\
        Answer repair & At most two follow-ups, with four model calls each \\
        Parallel evaluation & Five spans within one run \\
        Request idle timeout & 300 seconds \\
        \hline
    \end{tabular}
\end{table}

\subsubsection{Statistical Procedures and Cost Accounting}
\label{app:exp-stats}

\paragraph{Task outcomes.}
The paired bootstrap resamples 500 tasks with replacement, retaining each sampled task's four memory outcomes and four baseline outcomes~\citep{miller2024addingerrorbarsevals}. Each of 5{,}000 replicates computes the resolve-rate difference, and the 95\% interval uses the 2.5th and 97.5th percentiles. These intervals describe task variation conditional on the recorded sequences. Independent task resampling does not model dependence between tasks induced by evolving memory histories, which are not rerun. Across-run variation is summarized by the sample standard deviation of the four resolve rates.

Task gains and losses are set differences between resolved-task lists. For each memory configuration, we average over all 16 memory--baseline run pairs; baseline variation uses the 12 ordered pairs of distinct baseline runs.

\paragraph{Memory statistics.}
Counts sum over four runs, while profile shares average over all profiled spans with equal weight per span. Coverage uses all 500 episodes of each run, so an item is recurring when delivered in at least 50 episodes of that run. Token estimates include every delivery in the evaluated episodes, including restatements and spans without profiles.

\paragraph{Episode cost.}
\label{app:exp-cost-accounting}
Table~\ref{tab:episode-cost} reports agent and memory-system costs computed from recorded usage at the saved model rates, averaged over 2{,}000 episodes per configuration and excluding the judge. Accounting covers the 500 credited episodes of each run and their retained trajectory segments. Discarded attempts are excluded, and failed requests without reported usage contribute zero to this recorded-cost metric. Agent calls alone increase cost over the baseline by 8.8\% for MemGovern, 23.1\% for Mem0~(L+G), 23.4\% for Mem0~(L), and 37.8\% for TencentDB; recorded memory-system calls account for the remaining observed cost.

\subsection{Additional Results}
\label{app:exp-results}

The following analyses examine variation across runs, delayed use, delivery forms, and memory origins. They also report the coverage checks and computational resources required to obtain the accepted assessments.

\subsubsection{Memory Profiles, Uncertainty, and Coverage}
\label{app:exp-profiles}

All 15{,}271 judged spans received profiles, with no error or unjudged spans remaining. Table~\ref{tab:main-results} reports the pooled profiles, entry yields, uncertainty, episode coverage, and delivery volume. Each share uses all entries of its span, bounding corroboration and information gain by relevance, and benefit and harm by acted-on use. This common denominator also explains why low uncertainty shares can conceal substantial uncertainty among used entries. The mean unknown-effect share reaches at most 1.06\%, although 213 of 714 acted-on entries, or 29.8\%, have undetermined effects.

\begin{table}[t]
    \centering
    \small
    \vspace{-10pt}
    \caption{Memory profiles, entry yields, uncertainty, and coverage, pooled over four runs. Profile and uncertainty shares average over profiled spans, with all entries of each span in the denominator. Yields count beneficial or harmful entries per 100 profiled spans. Coverage uses all 2,000 episodes per configuration; recurring items occur in at least 50 episodes of a run. Delivered tokens include restatements and spans without profiles. Unknown novelty share is zero in every configuration.}
    \vspace{5pt}
    \label{tab:main-results}
    \begin{tabular}{@{}lrrrr@{}}
        \hline
        & MemGovern & Mem0 (L+G) & Mem0 (L) & TencentDB \\
        \hline
        \multicolumn{5}{@{}l}{\emph{Mean entry share (\%)}} \\
        Relevance & 15.08 & 1.42 & 1.43 & 28.48 \\
        Corroboration & 3.39 & 0.23 & 0.22 & 2.19 \\
        Information gain & 11.12 & 1.03 & 1.03 & 26.13 \\
        Acted on & 3.95 & 0.40 & 0.51 & 4.08 \\
        Benefit & 2.94 & 0.29 & 0.35 & 2.30 \\
        Harm & 0.36 & 0.01 & 0.01 & 0.03 \\
        \hline
        \multicolumn{5}{@{}l}{\emph{Entry yield (entries per 100 profiled spans)}} \\
        Benefit & 3.04 & 2.60 & 3.22 & 2.45 \\
        Harm & 0.36 & 0.06 & 0.06 & 0.03 \\
        \hline
        \multicolumn{5}{@{}l}{\emph{Mean uncertainty share (\%)}} \\
        Uncontradicted accuracy & 10.74 & 1.19 & 1.20 & 26.20 \\
        Unknown influence & 0.89 & 0.05 & 0.07 & 0.09 \\
        Unknown effect & 0.54 & 0.08 & 0.13 & 1.06 \\
        \hline
        \multicolumn{5}{@{}l}{\emph{Episode coverage (\%)}} \\
        Any profiled span & 73.10 & 97.60 & 97.45 & 99.80 \\
        Relevant entry & 27.30 & 11.95 & 12.40 & 99.80 \\
        \quad excluding recurring items & 27.30 & 11.75 & 12.20 & 14.30 \\
        Beneficial entry & 4.00 & 3.65 & 4.50 & 7.05 \\
        Harmful entry & 0.50 & 0.10 & 0.10 & 0.10 \\
        \hline
        \multicolumn{5}{@{}l}{\emph{Delivered-text cost and efficiency}} \\
        Delivered tokens (M) & 1.66 & 21.11 & 21.94 & 46.37 \\
        Beneficial entries / 1M tokens & 50.50 & 4.07 & 4.92 & 3.08 \\
        \hline
    \end{tabular}
\end{table}

\begin{figure}[t]
    \centering
    \includegraphics[width=0.9\linewidth]{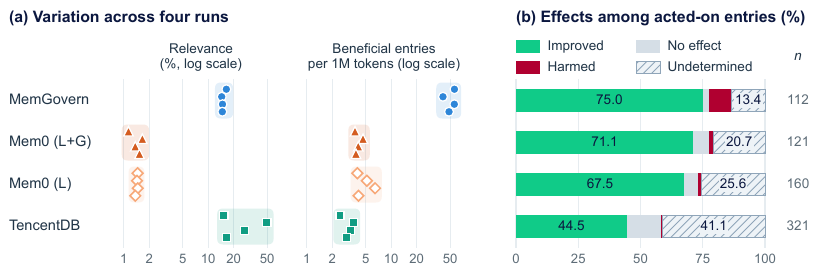}
    \caption{Run variation and effects conditional on use. (a) Markers show each of four runs per configuration on logarithmic scales; within a row, runs 1--4 are offset from top to bottom, and shading spans the four runs. Relevance is the mean entry share across profiled spans; token efficiency divides beneficial-entry counts by all estimated delivered tokens, including restatements and spans without profiles. (b) Effect counts are pooled across four runs and divided by the number of acted-on entries, $n$. Labels give improved and undetermined shares; every segment retains its proportional width.}
    \label{fig:appendix-diagnostics}
\end{figure}

\subsubsection{Per-Run Results}
\label{app:exp-per-run}

Figure~\ref{fig:appendix-diagnostics}a shows that the system differences persist across all four runs, with exact values in Table~\ref{tab:per-run}. Every Mem0 run has less than 1.7\% relevance, compared with more than 14\% for every MemGovern and TencentDB run. MemGovern exceeds 41 beneficial entries per million delivered tokens in every run, while the other configurations remain below 6.5. The ranges of the two Mem0 configurations overlap in every column, consistent with their observed use of the same local scopes.

TencentDB's relevance varies from 15.3\% to 49.6\%, partly because persona spans contain different numbers of scene-index entries. The persona entry is always relevant. It accompanies one scene in run 2 and three scenes in almost every such span of the other runs, giving these spans mean relevance of 98.2\% in run 2 and 29.8--50.7\% elsewhere. Across runs, the 1{,}996 persona spans constitute 34\% of profiled spans and contribute 18.0 percentage points to TencentDB's overall mean of 28.48\%. The remaining spans average 15.9\% relevance. Thus, the overall share reflects both the relevance of individual entries and their grouping into spans.

\begin{table}[t]
    \centering
    \small
    \vspace{-10pt}
    \caption{Results of each run. Resolved, relevance, acted on, benefit, and harm are percentages, with entry shares averaged over the profiled spans of the run. Efficiency counts beneficial entries per million delivered tokens, and spans counts profiled spans. The no-memory baseline resolved 75.0\%, 72.0\%, 73.0\%, and 71.8\% of the tasks in its four runs.}
    \vspace{5pt}
    \label{tab:per-run}
    \begin{tabular}{@{}lrrrrrrr@{}}
        \hline
        Run & Resolved & Relevance & Acted on & Benefit & Harm & Efficiency & Spans \\
        \hline
        \multicolumn{8}{@{}l}{\emph{MemGovern}} \\
        1 & 74.0 & 16.42 & 4.42 & 3.25 & 0.29 & 56.25 & 681 \\
        2 & 72.6 & 14.45 & 3.35 & 2.47 & 0.44 & 41.10 & 687 \\
        3 & 73.6 & 14.77 & 4.38 & 3.25 & 0.28 & 56.03 & 710 \\
        4 & 73.2 & 14.71 & 3.65 & 2.78 & 0.44 & 48.47 & 688 \\
        \hline
        \multicolumn{8}{@{}l}{\emph{Mem0 (L+G)}} \\
        1 & 73.4 & 1.13 & 0.35 & 0.26 & 0.02 & 3.73 & 830 \\
        2 & 71.8 & 1.66 & 0.49 & 0.36 & 0.00 & 4.64 & 856 \\
        3 & 73.2 & 1.36 & 0.38 & 0.27 & 0.02 & 4.07 & 817 \\
        4 & 72.4 & 1.53 & 0.38 & 0.27 & 0.00 & 3.82 & 808 \\
        \hline
        \multicolumn{8}{@{}l}{\emph{Mem0 (L)}} \\
        1 & 72.6 & 1.46 & 0.51 & 0.27 & 0.00 & 3.97 & 859 \\
        2 & 73.2 & 1.43 & 0.42 & 0.35 & 0.00 & 5.15 & 838 \\
        3 & 73.2 & 1.46 & 0.65 & 0.47 & 0.02 & 6.44 & 847 \\
        4 & 72.0 & 1.37 & 0.47 & 0.31 & 0.02 & 4.09 & 809 \\
        \hline
        \multicolumn{8}{@{}l}{\emph{TencentDB}} \\
        1 & 73.8 & 15.34 & 3.25 & 2.19 & 0.00 & 2.47 & 1{,}278 \\
        2 & 72.0 & 49.61 & 4.11 & 2.20 & 0.06 & 3.57 & 1{,}668 \\
        3 & 72.2 & 26.79 & 4.79 & 2.39 & 0.00 & 3.31 & 1{,}576 \\
        4 & 72.6 & 16.49 & 4.01 & 2.43 & 0.08 & 2.97 & 1{,}319 \\
        \hline
    \end{tabular}
\end{table}

\subsubsection{Use and Observed Effects}
\label{app:exp-use}

Conditioning on acted-on entries reveals uncertainty that the all-entry profile shares compress. Figure~\ref{fig:appendix-diagnostics}b shows undetermined effects for 13.4\% of MemGovern's acted-on entries, 20.7\% for Mem0~(L+G), 25.6\% for Mem0~(L), and 41.1\% for TencentDB. Improved effects account for 75.0\%, 71.1\%, 67.5\%, and 44.5\%, respectively. Table~\ref{tab:use-effects} retains the exact counts and separately reports episode outcomes. An episode containing both beneficial and harmful entries contributes to both episode groups.

Memory-shaped actions also often occur after the first consuming step. Among 714 acted-on entries, 51.8\% were first acted on at a later step and 11.1\% at least five steps later. Delayed use was most common in Mem0, whose task recall remains in the system message throughout the episode. Credit windows retain these actions until a later copy takes over or the episode ends.

\begin{table}[t]
    \centering
    \small
    \vspace{-10pt}
    \caption{Use and observed effects of judged entries, pooled over four runs. Unused relevant entries have influence \emph{none}, and novelty is judged against the agent's visible context. Delays compare the first anchored action with the first consuming step. Episode counts exclude the three episodes without a harness result; one MemGovern episode belongs to both effect groups.}
    \vspace{5pt}
    \label{tab:use-effects}
    \begin{tabular}{@{}lrrrr@{}}
        \hline
        & MemGovern & Mem0 (L+G) & Mem0 (L) & TencentDB \\
        \hline
        \multicolumn{5}{@{}l}{\emph{Relevant entries}} \\
        Count & 1{,}247 & 571 & 571 & 4{,}118 \\
        Unused (\%) & 51.32 & 79.68 & 77.06 & 75.06 \\
        Novel among unused (\%) & 58.91 & 49.45 & 48.18 & 89.52 \\
        \hline
        \multicolumn{5}{@{}l}{\emph{Acted-on entries}} \\
        Count & 112 & 121 & 160 & 321 \\
        Improved & 84 & 86 & 108 & 143 \\
        No effect & 3 & 8 & 9 & 44 \\
        Harmed & 10 & 2 & 2 & 2 \\
        Undetermined & 15 & 25 & 41 & 132 \\
        \hline
        \multicolumn{5}{@{}l}{\emph{Accuracy and relevance of entries with an effect (counts)}} \\
        Improved, contradicted & 8 & 1 & 0 & 1 \\
        Harmed, contradicted & 2 & 0 & 0 & 1 \\
        Harmed, irrelevant & 1 & 1 & 2 & 1 \\
        \hline
        \multicolumn{5}{@{}l}{\emph{First action after the first consuming step (\% of acted-on entries)}} \\
        Any later step & 34.82 & 74.38 & 66.88 & 41.74 \\
        Five or more steps later & 5.36 & 17.36 & 6.88 & 12.77 \\
        \hline
        \multicolumn{5}{@{}l}{\emph{Episode outcomes (episodes with at least one indicated effect)}} \\
        Beneficial use, passed & 61 & 55 & 66 & 112 \\
        Beneficial use, failed & 19 & 18 & 24 & 28 \\
        Harmful use, passed & 1 & 1 & 2 & 1 \\
        Harmful use, failed & 9 & 1 & 0 & 1 \\
        \hline
    \end{tabular}
\end{table}

\subsubsection{Task Outcomes and Memory Effects by Task Difficulty}
\label{app:exp-difficulty}

Beneficial use appears across the baseline-difficulty groups in Table~\ref{tab:difficulty}, reaching 3.4--5.0\% of memory episodes in each group. Changes in task outcome often lack a corresponding acted-on entry. Of 71 episodes resolving a task that all baseline runs failed, 7 contained acted-on use and 6 contained benefit. Of 87 episodes failing a task that all baseline runs resolved, 85 received memory, but only 7 contained acted-on use and 1 contained harm. These counts distinguish observed contributions within a trajectory from differences between final task outcomes.

\begin{table}[t]
    \centering
    \small
    \vspace{-10pt}
    \caption{Memory episodes of all four configurations, grouped by the number of baseline runs that resolved their task. The three episodes without a harness result are excluded.}
    \vspace{5pt}
    \label{tab:difficulty}
    \begin{tabular}{@{}lrrrrr@{}}
        \hline
        Baseline runs that resolved the task & 0 & 1 & 2 & 3 & 4 \\
        \hline
        Tasks & 104 & 22 & 13 & 33 & 328 \\
        Memory episodes & 1{,}664 & 352 & 208 & 527 & 5{,}246 \\
        \quad resolved (\%) & 4.27 & 27.84 & 44.71 & 77.42 & 98.34 \\
        Episodes with an acted-on entry & 144 & 36 & 17 & 36 & 411 \\
        Episodes with a beneficial entry & 79 & 12 & 7 & 25 & 260 \\
        \quad share of memory episodes (\%) & 4.75 & 3.41 & 3.37 & 4.74 & 4.96 \\
        \quad resolved (\%) & 7.59 & 25.00 & 71.43 & 88.00 & 99.23 \\
        Episodes with a harmful entry & 5 & 4 & 0 & 1 & 6 \\
        \hline
    \end{tabular}
\end{table}

\subsubsection{Delivery Forms and Repetition}
\label{app:exp-forms}

Full content supplies most beneficial entries despite occupying a small share of delivered text, as Table~\ref{tab:delivered-forms} shows. Forms are identified from annotation provenance for MemGovern and TencentDB and from placement slots for Mem0. Token shares include all deliveries, allocating each span's estimate among entries in proportion to exposed characters. Entry rates use accepted assessments, while automatic neutral copies retain token cost without an assessment.

\begin{table}[t]
    \centering
    \small
    \vspace{-10pt}
    \caption{Delivery volume and entry outcomes by form, pooled over four runs and combining both Mem0 configurations. H denotes host-scheduled delivery and A agent-directed delivery. Token and beneficial-entry shares use each system's totals, including 84, 194, and 143 beneficial entries for MemGovern, Mem0, and TencentDB. Entry rates use judged entries of each form. Copies with empty credit windows retain their token volume without receiving an assessment. Omitted empty and automatic-copy forms each contribute less than 0.1\% of tokens and remain in the denominators.}
    \vspace{5pt}
    \label{tab:delivered-forms}
    \begin{tabular}{@{}llrrrrr@{}}
        \hline
        & & \multicolumn{2}{c}{Share of system (\%)} & \multicolumn{2}{c}{Entry rate (\%)} & \\
        \cline{3-4}\cline{5-6}
        Delivered form & Route & Tokens & Beneficial & Relevant & Acted on & Harmful (n) \\
        \hline
        \multicolumn{7}{@{}l}{\emph{MemGovern}} \\
        Search preview & A & 87.52 & 3.57 & 4.48 & 0.01 & 0 \\
        Full card & A & 12.48 & 96.43 & 45.77 & 15.35 & 10 \\
        \hline
        \multicolumn{7}{@{}l}{\emph{Mem0}} \\
        Task recall & H & 12.96 & 75.77 & 2.00 & 0.88 & 4 \\
        Search result & A & 18.95 & 24.23 & 0.97 & 0.11 & 0 \\
        Copy of visible entries & H, A & 68.08 & -- & -- & -- & -- \\
        \hline
        \multicolumn{7}{@{}l}{\emph{TencentDB}} \\
        Skill catalog & H & 59.54 & 0.00 & 0.34 & 0.02 & 0 \\
        Persona & H & 26.05 & 3.50 & 100.00 & 0.45 & 0 \\
        Scene index & H, A & 9.11 & 0.00 & 24.57 & 1.58 & 0 \\
        Skill search preview & A & 2.10 & 4.20 & 1.69 & 0.10 & 0 \\
        Full skill & A & 1.99 & 86.01 & 21.25 & 20.92 & 1 \\
        Scene content & A & 1.15 & 6.29 & 96.63 & 5.53 & 1 \\
        \hline
    \end{tabular}
\end{table}

Both Mem0 configurations use the same local scopes and are combined in the table. Agent-directed searches supply 75.4\% of judged entries, with 0.97\% relevance, while task recall supplies 147 of 194 beneficial entries. Repeated copies that add nothing to visible content form 3{,}208 spans with empty credit windows and consume 68.1\% of delivered tokens. Of that volume, 99.6\% appears in the system-message suffix and the remainder in search results.

TencentDB's persona, scene index, and skill catalog account for 94.7\% of delivered tokens and 5 of 143 beneficial entries. Full skills, scene content, and skill-search previews supply the other 138. Catalogs and previews also guide access to fuller content through 1{,}500 selections, comprising 742 skill-catalog entries, 412 scene-index entries, and 346 skill-search previews. Almost all scene-index entries are host-scheduled, with only 5 of 4{,}986 coming from agent-directed listings.

Recurring items account for 88.6\% of TencentDB's relevant entries, mainly through the persona block. The block is relevant in all 1{,}996 episodes receiving it, acted on in 9, and beneficial in 5. Excluding recurring items from the coverage calculation reduces relevance coverage from 99.8\% to 14.3\%. This calculation identifies which items account for broad relevance coverage in the recorded runs.

Restatements account for 45{,}279 of TencentDB's 51{,}242 deliveries and 43{,}077 of Mem0's 55{,}051. Across configurations, neutral spans receive 97.5--99.1\% of delivered tokens. This category includes automatic copies, selections, and references, so its volume combines unused text with text that supports access to separately assessed content.

\subsubsection{Memory Histories}
\label{app:exp-histories}

Memory histories connect later benefit to experience from earlier tasks. Table~\ref{tab:memory-origins} traces all 194 beneficial Mem0 entries and 138 of TencentDB's 143 to earlier episodes in the same run. Mem0's traced entries all originate in the same repository as their use, while 13 TencentDB entries originate in another repository. Median gaps of 19--53 tasks show that useful items persist beyond neighboring episodes. These are counts of beneficial assessments, so a stored item can contribute more than once.

Failed source episodes produced 73 of the 331 traced entries with known source-task outcomes, or 22.1\%. One additional traced TencentDB entry has a source episode without a harness result and is excluded from this denominator. Thus, a failed attempt can supply useful guidance to a later task even when it does not solve its own task.

\begin{table}[t]
    \centering
    \small
    \vspace{-10pt}
    \caption{Origins of beneficial entries, pooled over four runs. Traced origins are earlier episodes in the same run. Source-outcome counts exclude one TencentDB entry whose creation episode has no harness result. Gaps count task positions from creation to beneficial use among traced entries.}
    \vspace{5pt}
    \label{tab:memory-origins}
    \begin{tabular}{@{}lrrr@{}}
        \hline
        & Mem0 (L+G) & Mem0 (L) & TencentDB \\
        \hline
        Beneficial entries & 86 & 108 & 143 \\
        Traced origins & 86 & 108 & 138 \\
        \quad with known source outcome & 86 & 108 & 137 \\
        \quad from failed source episodes & 23 & 26 & 24 \\
        \quad from another repository & 0 & 0 & 13 \\
        Median gap in tasks & 40 & 53 & 19 \\
        \hline
    \end{tabular}
\end{table}

MemGovern's fixed corpus also permits comparison of repeated uses. Its 84 beneficial entries came from 49 cards, including 21 that helped in at least two runs. Five cards produced its 10 harmful entries. A Django card harmed \texttt{django-14170} in all four runs, and a SymPy card harmed \texttt{sympy-23950} in three; every acted-on use of these two cards was harmful. Two other cards each helped the same task twice and harmed it once across runs, for \texttt{sympy-18698} and \texttt{sphinx-8638}. The fifth harmed \texttt{sympy-12419} once.

\subsubsection{Benchmark Overlap in MemGovern's Corpus}
\label{app:memgovern-overlap}

An external repair corpus introduces a risk of \emph{retrieval-time benchmark leakage}: the agent may receive solution details from the same problem it is being evaluated on. In MemGovern, this overlap arises through cards derived from the benchmark problems' own issue discussions and repair attempts. For example, the card retrieved on \texttt{sphinx-8475} comes from \href{https://github.com/sphinx-doc/sphinx/pull/8131}{Sphinx PR~\#8131}, whose fix was subsequently cherry-picked into the \href{https://github.com/sphinx-doc/sphinx/pull/8475}{benchmark PR}. Its advice to catch \texttt{TooManyRedirects} and fall back to a GET request matches the gold repair.

Matching card identifiers to GitHub records also links beneficial use to earlier attempts at the benchmark problems. The card used on \texttt{sympy-13974} comes from \href{https://github.com/sympy/sympy/pull/13784}{SymPy PR~\#13784}, an alternative attempt to solve the same issue. The card used on \texttt{pytest-5840} comes from \href{https://github.com/pytest-dev/pytest/pull/5823}{Pytest PR~\#5823}, which the benchmark fix superseded. The \texttt{sympy-20438} card in Figure~\ref{fig:case-studies} comes from \href{https://github.com/sympy/sympy/pull/16764}{SymPy PR~\#16764}, whose discussion contains the comment reproduced in the benchmark problem statement. These three cards contributed 7 of the 84 beneficial entry assessments (8.3\%).

Source overlap can also preserve superseded guidance. The card judged harmful on \texttt{django-14170} in all four runs summarizes \href{https://github.com/django/django/pull/14157}{PR~\#14157}, which proposed removing the ISO-year lookup optimization. The accepted \href{https://github.com/django/django/pull/14170}{benchmark fix} instead retained the optimization and corrected its date bounds. MemGovern's reported relevance and token efficiency therefore include guidance from the benchmark problems' own development histories. This targeted review establishes specific overlaps; their prevalence across the full corpus remains unmeasured.

\subsubsection{Judge Execution and Resource Use}
\label{app:exp-judge-runs}

\paragraph{Follow-ups.}
Evidence checks triggered follow-ups for 949 of 15{,}271 judged spans. Rejection reasons overlap, including incomplete reads in 692 spans, invalid JSON in 173, other answer-contract violations in 96, anchors on unread steps in 55, and structural errors in 3. In 627 of the incomplete-read cases, the judge had assigned \emph{none}, \emph{referenced}, or \emph{selected} before reading the full credit window. These follow-ups enforced the evidence access required for claims about non-use or limited use.

Among the 809 follow-up spans with a parseable first answer, 17 changed at least one signal before acceptance, including influence in 11, accuracy in 5, relevance in 3, and novelty in 1. Within the 687 parseable cases rejected for coverage, 8 changed a signal and 5 changed influence. Thus, most repairs expanded evidence access without changing the signal labels. Nineteen spans required a forced final answer, and all ultimately passed validation.

\paragraph{Cost.}
Judging required 77{,}329 model calls, averaging 5.06 per judged span, as detailed in Table~\ref{tab:judge-cost}. The prompt cache served 74.5\% of input tokens. TencentDB required the most follow-ups and calls per span. The batch runner evaluated one run at a time with five spans in parallel, totaling 31.5 hours across the recorded evaluation segments.

\begin{table}[t]
    \centering
    \small
    \vspace{-10pt}
    \caption{Evaluation units, span triage, and judge usage, pooled over four runs. Spans split into judged, automatic neutral, and empty; triage counts include automatic neutral spans and exclude empty spans. Restatements are later deliveries within a span. Input tokens are split into uncached and cached tokens as reported by the provider.}
    \vspace{5pt}
    \label{tab:judge-cost}
    \begin{tabular}{@{}lrrrrr@{}}
        \hline
        & MemGovern & Mem0 (L+G) & Mem0 (L) & TencentDB & Total \\
        \hline
        \multicolumn{6}{@{}l}{\emph{Evaluation units}} \\
        Deliveries & 2{,}766 & 27{,}381 & 27{,}670 & 51{,}242 & 109{,}059 \\
        \quad restatements (\%) & 0.00 & 78.11 & 78.39 & 88.36 & 81.02 \\
        Spans & 2{,}766 & 5{,}994 & 5{,}980 & 5{,}963 & 20{,}703 \\
        \quad automatic neutral & 0 & 1{,}582 & 1{,}626 & 3 & 3{,}211 \\
        \quad empty & 0 & 1{,}101 & 1{,}001 & 119 & 2{,}221 \\
        Judged spans & 2{,}766 & 3{,}311 & 3{,}353 & 5{,}841 & 15{,}271 \\
        Judged entries & 21{,}270 & 46{,}553 & 47{,}071 & 57{,}300 & 172{,}194 \\
        \hline
        \multicolumn{6}{@{}l}{\emph{Span triage (counts)}} \\
        Positive & 84 & 73 & 90 & 142 & 389 \\
        Negative & 10 & 2 & 2 & 2 & 16 \\
        Unresolved & 15 & 22 & 31 & 130 & 198 \\
        Neutral & 2{,}657 & 4{,}796 & 4{,}856 & 5{,}570 & 17{,}879 \\
        \hline
        \multicolumn{6}{@{}l}{\emph{Judge usage}} \\
        Spans with a follow-up & 61 & 161 & 141 & 586 & 949 \\
        Forced final answers & 0 & 2 & 3 & 14 & 19 \\
        Model calls & 11{,}599 & 16{,}483 & 16{,}770 & 32{,}477 & 77{,}329 \\
        \quad per judged span & 4.19 & 4.98 & 5.00 & 5.56 & 5.06 \\
        Uncached input (M) & 75.3 & 136.4 & 138.0 & 234.6 & 584.3 \\
        Cached input (M) & 190.1 & 364.6 & 368.1 & 787.0 & 1{,}709.8 \\
        Output (M) & 16.1 & 25.4 & 26.1 & 42.8 & 110.4 \\
        Wall time (h) & 4.5 & 6.6 & 7.1 & 13.4 & 31.5 \\
        \hline
    \end{tabular}
\end{table}

\end{document}